\documentclass[lettersize,journal, twoside]{IEEEtran}
\usepackage{amsmath,amsfonts}
\usepackage{algorithm}
\usepackage[noend]{algpseudocode}
\usepackage{array}
\usepackage[caption=false,font=normalsize,labelfont=sf,textfont=sf]{subfig}
\usepackage{textcomp}
\usepackage{stfloats}
\usepackage{url}
\usepackage{verbatim}
\usepackage{graphicx}
\usepackage{cite}
\usepackage{xcolor}
\begin{document}

\title{Safe Start Regions for Medical \\Steerable Needle Automation}

\author{Janine Hoelscher,~\IEEEmembership{Student Member,~IEEE,}
    Inbar Fried,~\IEEEmembership{Student Member,~IEEE,}
    Spiros Tsalikis,
    Jason Akulian,
    Robert J. Webster III,~\IEEEmembership{Senior Member,~IEEE,} and
    Ron Alterovitz,~\IEEEmembership{Senior Member,~IEEE}
        %
\thanks{This research was supported by the U.S. National Institutes of Health (NIH) under award R01EB024864 and the National Science Foundation (NSF) under awards 2008475 and 2038855.}%
 \thanks{J. Hoelscher, I. Fried, S. Tsalikis, and R. Alterovitz are with the Department of Computer Science, University of North Carolina at Chapel Hill, Chapel Hill, NC 27599, USA. (email: \{jhoelsch, ifried01, spiros, ron\}@cs.unc.edu)}%
\thanks{J. A. Akulian is with the Division of Pulmonary Diseases and Critical Care Medicine at the University of North Carolina at Chapel Hill, NC 27599, USA. (email: jason\_akulian@med.unc.edu}%
\thanks{R. J. Webster III is with the Department of Mechanical Engineering, Vanderbilt University, Nashville, TN 37235, USA. (email: robert.webster@vanderbilt.edu)}%
}

\markboth{Preprint}
{\textit{Hoelscher \MakeLowercase{et al.}}: Safe Start Poses for Medical Steerable Needle Automation}

\IEEEpubid{0000--0000/00\$00.00~\copyright~2024 IEEE}
\maketitle

\begin{abstract}

Steerable needles are minimally invasive devices that enable novel medical procedures by following curved paths to avoid critical anatomical obstacles. Planning algorithms can be used to find a steerable needle motion plan to a target. Deployment typically consists of a physician manually inserting the steerable needle into tissue at the motion plan’s start pose and handing off control to a robot, which then autonomously steers it to the target along the plan. The handoff between human and robot is critical for procedure success, as even small deviations from the start pose change the steerable needle's workspace and there is no guarantee that the target will still be reachable. We introduce a metric that evaluates the robustness to such start pose deviations. When measuring this robustness to deviations, we consider the tradeoff between being robust to changes in position versus changes in orientation. We evaluate our metric through simulation in an abstract, a liver, and a lung planning scenario. Our evaluation shows that our metric can be combined with different motion planners and that it efficiently determines large, safe start regions.

\end{abstract}

\section{Introduction}
Steerable needles are medical instruments that have the potential to enable novel, minimally invasive procedures. Their ability to follow curved paths allows them to circumvent critical anatomical obstacles such as large blood vessels or bones, which opens up new options for accessing previously hard-to-reach sites in the body. Steerable needles have been demonstrated to be versatile in supporting different types of procedures (e.g., tissue ablation \cite{adebar2015methods}, localized drug delivery \cite{Secoli2022_PLOSOne}, and diagnostic biopsies \cite{kuntz2023autonomous}). Recently, they have been successfully used in in-vivo experiments in the brain \cite{Secoli2022_PLOSOne} and in the lungs \cite{kuntz2023autonomous}.

Steerable needles are not intuitive for humans to control manually due to their non-holonomic and curvature constraints \cite{majewicz2013cartesian}. As a result, a steerable needle procedure often consists of a physician manually positioning and orienting a steerable needle at a precomputed start pose and handing off control to a robot, which then autonomously steers the steerable needle to the target. The \emph{handoff} between the physician and the autonomous robot is a critical step for the procedure's success.

\begin{figure}[t]
\vspace{5pt}
 \centering
 \includegraphics[width=1.0\columnwidth]{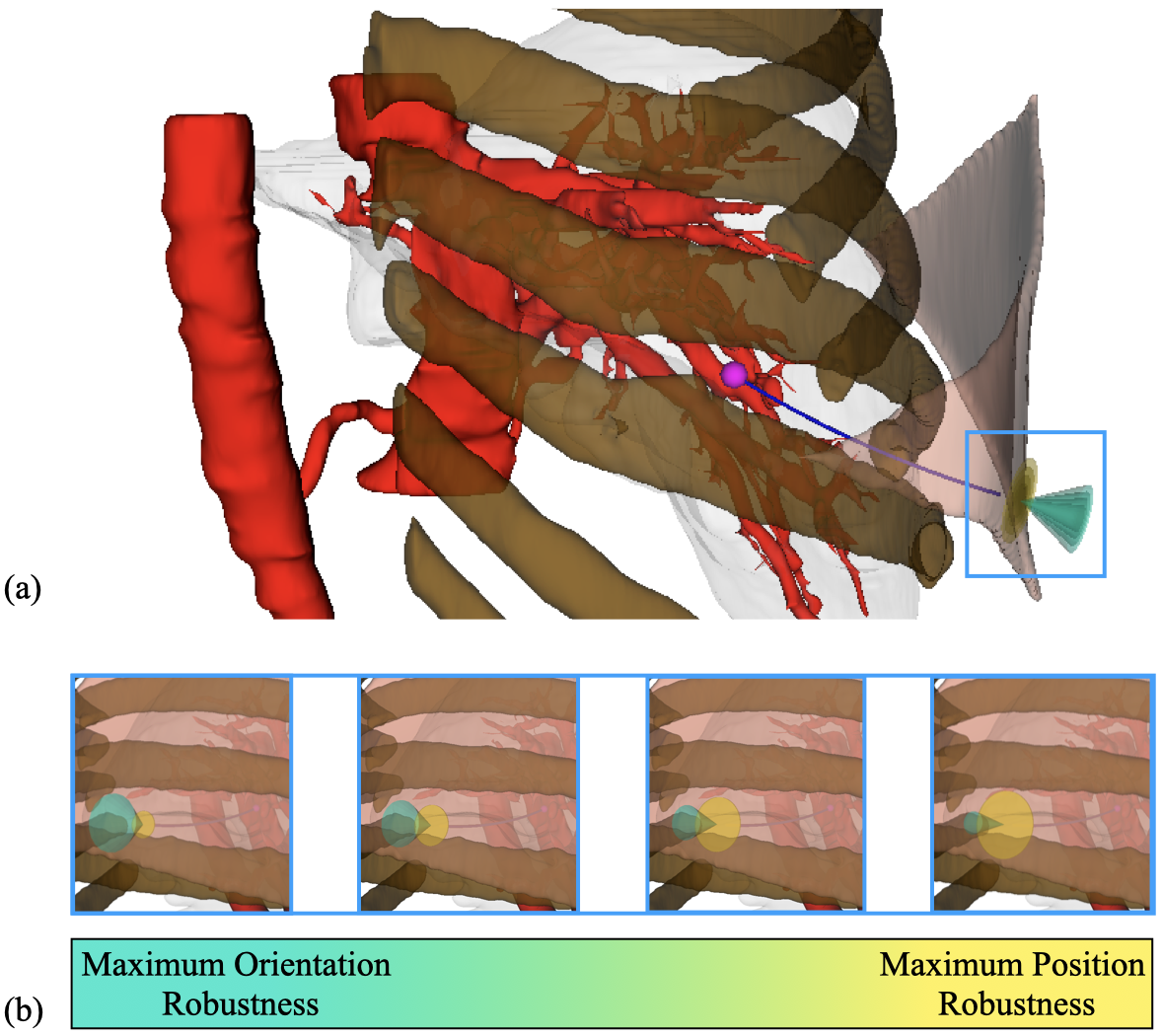}
 \caption{
 (a) We demonstrate our method in a liver biopsy scenario where a steerable needle is inserted from the skin (tan) into liver tissue (grey) and is steered towards a target (pink) representing a suspicious nodule while avoiding large blood vessels (red) and the ribs (brown). A physician first manually inserts the steerable needle into the insertion surface, then the physician hands off control to an automated robot which steers the steerable needle through tissue and to the target.
 Our method determines a safe start region around the start pose of a nominal motion plan (dark blue) on an insertion surface (tan). Poses in this safe start region can be both reached by the physician, and a collision-free motion plan to the target can be found. 
 We determine the size of the safe start region in both position (yellow) and orientation (cyan). 
 (b) There is a tradeoff between maximizing robustness in position and in orientation, and our method allows the user to choose this tradeoff according to their preferences and the specific insertion scenario.
 }
 \label{fig:first}
 \end{figure}

\IEEEpubidadjcol 

The position and orientation of the steerable needle when the physician hands off the needle to the robot has a significant impact on whether the robot will be able to autonomously steer the needle to the target. 
As the steerable needle's maximum curvature in tissue has limits, some targets in obstacle-cluttered environments may not be accessible from some initial start poses.  
To address this challenge, we can use motion planning to identify the position and orientation of a favorable handoff pose \cite{hoelscher2021backward} and a path on which to safely steer the needle from the handoff pose to the target while avoiding obstacles \cite{patil2014needle, webster2006nonholonomic}. Here, reasoning about both the steerable needle's start \emph{position} and its start \emph{orientation} are crucial to guarantee target reachability. Even a small deviation from a planned start pose in either position or orientation could result in the target being unreachable.

To help physicians in positioning and orienting a steerable needle for a handoff to automated steering, we developed a method to efficiently calculate the amount of deviation in position and orientation that is allowable to guarantee that the target will be reachable. 
Our method computes a set of safe start poses around the nominal start pose, for which we can guarantee that an obstacle-free motion plan to the target can be found. We use the size of this safe start region as a metric for evaluating the plan's start pose robustness. 
Here, a \emph{tradeoff} between position and orientation robustness has to be considered: if a physician desires high robustness to changes in position, meaning a large start area, the overall angular tolerance associated with all positions in this area will be smaller. Conversely, a desire for a large angular tolerance will result in a small number of corresponding positions from which there is guaranteed target reachability and, therefore, a smaller position robustness. 
Our new method is able to compute both safe start positions and orientations as well as provide a user-specific input parameter for the tradeoff between maximizing the two as visualized in Fig.~\ref{fig:first}.

Our new method is not tied to any particular steerable needle planning algorithm and can be applied to any motion plan fulfilling a few conditions outlined in Section \ref{sec:problem} (e.g., the motion plan has to begin on an insertion surface). It can also be used as a start pose robustness metric to select from multiple motion plans that a motion planner computed. We applied this metric to plans created by two different steerable needle motion planners: the steerable needle RRT \cite{patil2014needle} and an AO-RRT \cite{hauser2016asymptotically} adapted to steerable needles. Our method can take into account additional orientation restrictions on the start region, which can be imposed by hardware limitations or anatomical obstructions, making our method suitable for deployment in various planning scenarios for different types of procedures. We demonstrate this versatility in three different planning scenarios.
 First, we created an abstract scenario consisting of a flat start region and spherical obstacles for systematic testing. Second, in a transoral lung access scenario, we determine airway areas that can be reached by a bronchoscope, and we apply our start pose robustness metric to motion plans from the surface of these airways to remote nodules in the lung, simulating a biopsy procedure. Third, in a percutaneous liver biopsy scenario, we determine safe start regions on the patient's skin from which areas in the liver can be reached safely while avoiding ribs in close proximity to the liver.
We also show that our method is efficient and finds larger safe start regions significantly faster than a Monte Carlo sampling-based comparison method. 
\section{Related Work}
Robotic steerable needles are a promising approach to reaching sites in the body not safely reachable with conventional instruments \cite{cowan2011robotic}. One type of steerable needle has a beveled tip, which applies an asymmetric force on tissue when it is inserted, causing it to curve in the direction of the bevel \cite{webster2006nonholonomic}.
The needle's curvature is limited by a minimum radius of curvature \cite{webster2006nonholonomic, rox2020laser}, which it achieves when it is inserted into tissue without any axial rotation. Control strategies varying the insertion and axial rotation velocities can be used to achieve different needle curvatures and to let the needle follow complex paths in 3D \cite{minhas2007modeling, rucker2013sliding, ertop2020steerable}. The curvature constraint significantly limits the steerable needle's reachable workspace \cite{fried2021design}, which makes it difficult for human operators to reason about deployment paths in 3D. Motion planning algorithms can help with this problem.

\subsection{Steerable Needle Motion Planning} 
Steerable needle motion planners can determine plans that comply with the needle's hardware constraints and avoid collisions with critical obstacles. A variety of different motion planning techniques have been applied to steerable needles. 
Optimization-based planning techniques create an initial candidate plan and then improve it according to a cost function \cite{duindam2010three, bano2012smooth, segato2021optimized, li2014combination}. 
Sampling-based motion planners are an efficient way to find plans in high-dimensional spaces and have been successfully applied to steerable needles \cite{caborni2012risk, xu2008motion, sun2015stochastic}. One popular sampling-based technique is the Rapidly Exploring Random Tree (RRT) \cite{lavalle1998rapidly} algorithm, which can take the needle's hardware constraints into account \cite{patil2010interactive, favaro2021evolutionary}.

Search-based planning methods provide a systematic way to search the reachable workspace and to find motion plans. Steerable needle planners include a fractal tree approach ~\cite{liu2016fast, pinzi2019adaptive} and a resolution-complete search tree strategy \cite{fu2023certifiable}.
Most recently, deep learning methods have been applied to steerable needle motion planning \cite{segato2021inverse}.
While the steerable needle motion planners listed so far are used for planning before the procedure, there also exist techniques for fast re-planning during deployment, which can be used to handle uncertainty or deviations \cite{sun2015replanning,pinzi2021replanning, patil2014needle}.

Most motion planning problems assume a fixed start pose that coincides with the robot's current configuration. However, during a procedure, steerable needles are deployed from an insertion surface, which results in multiple potential start poses. Such a start pose can be determined by letting the physician choose \cite{segato2022hybrid}, by random sampling \cite{kuntz2015motion}, or by computing the start pose that results in the best motion plan
  \cite{pinzi2019adaptive, alterovitz2008motion}.
Our previous work \cite{hoelscher2021backward} inverts the planning problem and plans backward from a fixed target position towards a start region.
We use this backward planning strategy to create nominal motion plans in this work. However, our metric is agnostic to the planning algorithm, and other steerable needle motion planners could be considered instead. 

In this work, we are interested in the robustness to deviations in the start pose of a needle plan. 
A related concept is steerable needle motion planning considering uncertainty during robot deployment \cite{park2005diffusion, alterovitz2008motion, van2012motion, sun2015stochastic}. 
Additional needle planning methods explicitly consider deformations and thereby mitigate the need to handle uncertainty \cite{alterovitz2005planning, patil2011motion, segato2021position}.

\subsection{Motion Planning Metrics}
A fundamental property of most motion planning strategies is evaluating plans using a cost metric to facilitate the selection of the best plan.  
Literature provides a large variety of metrics applied to steerable needle motion plans aiming to ensure procedure success and minimize patient risk.

A popular metric for steerable needle motion plans is path length, as shorter plans have less potential for risk during deployment \cite{duindam2008screw, burrows2015smooth, patil2014needle, fu2023certifiable}.
Distance to obstacles along the plan is another common metric that aims to reduce patient risk \cite{patil2014needle, kuntz2015motion, fu2023certifiable, duindam2008screw}. A variant of this strategy uses a cost map representing less critical obstacles that should be avoided as much as possible \cite{fu2018safe, caborni2012risk}.
Another class of metrics aims to evaluate the difficulty of accurately steering a steerable needle along a plan. The amount of curvature along a plan measures how much potential for recovery there is should the needle deviate from its plan \cite{pinzi2019adaptive, favaro2021evolutionary}, how much control effort is necessary for the needle to follow said plan \cite{pinzi2019adaptive, duindam2008screw}, and how much sheering through tissue could occur \cite{bentley2023safer}.
A related concept is to evaluate motion plans according to their robustness to deviations during the procedure. Strategies include evaluating plans according to their robustness to control error \cite{favaro2018uncertainty, alterovitz2008motion} or based on the distance of a path to the boundaries of the needle's reachable workspace \cite{pinzi2021replanning}.
Our previous work \cite{hoelscher2022metric} chooses a start position for a motion plan on an insertion surface based on its robustness to deviations in its start position during initial deployment. In this work, we introduce a new approach that considers start pose robustness not only in positions but also in orientations.

As all of the above metrics aim to ensure procedure success, choosing which one is most relevant for a particular application can be difficult. A popular strategy is to select multiple metrics and to calculate a weighted sum of their scores \cite{duindam2008screw, caborni2012risk, sun2015replanning, hong20193d, pinzi2021computer, favaro2021evolutionary}. Segato et al.~customize the planning process by letting users choose the weights of multiple metrics \cite{segato2019automated} or by learning a rewards function based on expert preferences \cite{segato2022hybrid}.

Beyond steerable needles, port placement for other minimally invasive procedures shares the challenge of choosing a safe start pose that ensures the reachability of the target. The tools used in these procedures only allow for straight insertions, which reduces planning effort.
Evaluation metrics for single ports include the overlap of the tool's reachable workspace with the surgical site \cite{feng2017pose}, the optimal tool angle at the surgical site \cite{cannon2003port}, distance to critical obstacles \cite{wankhede2019heuristic}, and robot dexterity at the target location \cite{adhami2000planning}.
In multi-port procedures, the relationship between port locations is critical \cite{hayashi2017optimal, sun2007port}. Furthermore, physician preference and visibility can be taken into account \cite{sun2007port}.

 Outside of medical applications, metrics evaluating handoffs between humans and robots can be found. These metrics evaluate if human users can reach a potential handoff pose \cite{vahrenkamp2016workspace} and how fast they can reach it \cite{ mainprice2012sharing}. 
Walker et al.~\cite{walker2015robot} designed a metric measuring the predictability of the human's behavior, which allows the robot to slow down if unexpected human movement is detected.
 However, there exists no method that evaluates handoff positions evaluating the robot's reachable workspace and its robustness to deviations in both position and orientation.

 \subsection{Dubins Paths}
 In this work, we introduce a method that efficiently computes orientation ranges from which the target can be reached by using Dubins Paths.
 A Dubins path is the shortest path between two poses (position and orientation) in 2D for a vehicle with non-holonomic constraints, i.e., a minimum radius of curvature of its path \cite{dubins1957curves}. There are six combinations of minimum curvature segments, 
 one of which is guaranteed to be the shortest path.  
 Although multiple methods for using Dubins paths in 3D have been proposed \cite{cai2017task,hota2010optimal,hota2014optimal,elbanhawi2014randomised,chitsaz2007time}, no general analytical solution for the 3D case has been found so far. 
 Dubins paths have been used to find the shortest paths for steerable needles in 2D \cite{alterovitz2008motion}. In 3D steerable needle planning, Dubins paths are combined with numerical techniques \cite{duindam2010three} or are used within an RRT planner \cite{fauser2018planning}. In this work, we use Dubins paths to determine the reachable workspace of a steerable needle based on its start pose.
\section{Problem Definition}
\label{sec:problem}

We first formally define the steerable needle model and general steerable needle motion planning problem and then formalize a metric for steerable needle motion plans. The metric evaluates the robustness of a motion plan's start pose to deviation in position and orientation during a handoff from a physician to an autonomous robot.

The steerable needle has a beveled tip and curves in the direction of the bevel when being inserted into tissue, as the tissue imposes a reaction force on the bevel \cite{webster2006nonholonomic}. Its reachable workspace is limited by the minimum radius of curvature $r_\text{min}$ it can achieve. We use a 3D unicycle model \cite{park2005diffusion, webster2006nonholonomic} to model the steerable needle's kinematics, assuming that the needle follows a constant curvature arc in the plane perpendicular to its bevel when inserted. Rotating the needle around its axis changes the direction of the bevel and therefore the steering direction. Duty cycling can achieve steering along arcs with larger radii of curvature. The needle's reachable workspace is further limited by the needle's diameter $d_\text{needle}$ (to avoid close-by obstacles) and its maximum length $l_\text{needle}$. Furthermore, the needle's overall curvature cannot exceed $\pi/2$ to avoid sheering through tissue or buckling \cite{patil2014needle}.

\begin{figure}[t]
 \centering
 \includegraphics[width=1.0\columnwidth]{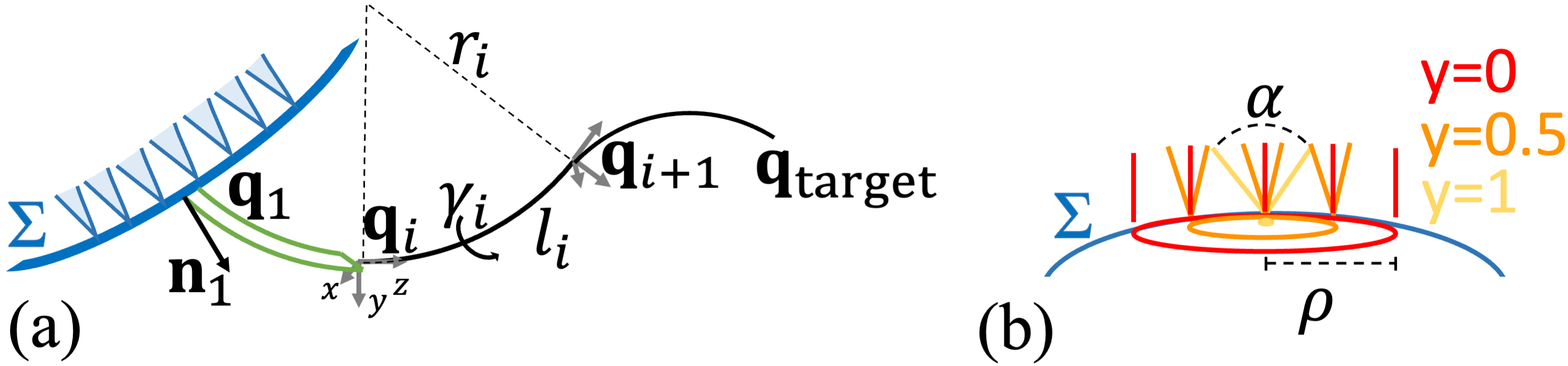}
 \caption{(a) The steerable needle (green) is deployed from the start region $\Sigma$ at configuration $\mathbf{q}_1 \in \Sigma$ and aims to reach $\mathbf{p}_\text{target}$. A needle plan (black) consists of constant curvature arcs characterized by an insertion length $l_i$, a radius of curvature $r_i$, and a rotation angle $\gamma_i$ that connect 3D poses $\mathbf{q}_i$. (b) Start pose robustness metric $R(\Pi, \Sigma,y)$ computes maximum allowable deviations depending on tradeoff parameter $y$, balancing robustness in position ($\rho$) and orientation ($\alpha$). 
 }
 \label{fig:definition}
 \end{figure}

We denote a steerable needle tip configuration as $\mathbf{q} \in \mathcal{SE}(3) = \mathbb{R}^3 \times \mathcal{SO}(3)$, expressing its position $\mathbf{p} \in \mathbb{R}^3$ and its orientation 
$\mathbf{R}\in \mathcal{SO}(3)$. Following the kinematics model described above, we model a section of a steerable needle motion plan as a constant curvature arc $\mathbf{a}_i = \{\mathbf{q}_i, r_i, l_i, \gamma_i\}$, with radius $r_i$, length $l_i$, and axial rotation $\gamma_i$, connecting configuration $\mathbf{q}_{i}$ to subsequent configuration $\mathbf{q}_{i+1}$, as visualized in Fig.~\ref{fig:definition} (a). We model deployment along a plan segment such that the steerable needle moves in direction $\mathbf{R}^z_i$ and curves into direction $-\mathbf{R}^y_i$.
A steerable needle motion plan is an ordered list of such constant curvature arc plan sections $\Pi  = [\mathbf{a}_{1}, \dots, \mathbf{a}_i, \dots \mathbf{a}_{n}]$ of $n \in \mathbb{N}$ plan sections beginning at start pose $\mathbf{q}_1$.
Assuming follow-the-leader deployment \cite{choset1999follow}, the steerable needle's configuration is defined by the consecutive configurations of its tip following a motion plan. 

The motion planning scenario for the steerable needle consists of a 3D workspace $\mathcal{W}\subset \mathbb{R}^3$ containing critical obstacles $\mathcal{O} \subset \mathcal{W}$ that need to be avoided as well as a target $\mathbf{p}_\text{target} \in \mathcal{W}$ that has to be reached.
The physician initially positions the needle on an insertion surface in $\mathcal{W}$. We define the set of all start configurations the physician can reach on this surface as start region $\Sigma \in \mathcal{SE}(3)$. 
The goal of the motion planner is to find a steerable needle motion plan $\Pi$ starting at $\mathbf{q}_1 \in \Sigma$  and connecting to $\mathbf{p}_{n+1} = \mathbf{p}_\text{target}$ that avoids collisions with obstacles $\mathcal{O} \subset \mathcal{W}$.

We evaluate a motion plan by quantifying its robustness to small deviations to its start pose in both position and orientation.
Our method computes $\mathbf{Q}_1 \subset \Sigma$, a safe start region defined by a set of start configurations surrounding the nominal start pose $\mathbf{q}_1$. These configurations are both achievable by the physician and have a collision-free motion plan to the target.
We introduce a start pose robustness metric $R(\Pi, \Sigma, y)\mapsto (\rho, \alpha)$, $y \in [0,1]$, $\rho \in \mathbb{R}^+$, $\alpha \in [0,\pi]$, that measures the size of $\mathbf{Q}_1$ and thereby expresses the tolerance to deviations in the start configuration for a nominal motion plan $\Pi$. We aim to optimize robustness in both position and orientation. 
However, if we increase $\mathbf{Q}_1$ to include more positions, the intersection of all orientation ranges at each of these positions will be smaller. Therefore, a larger tolerance to changes in position ($\rho$) results in a smaller tolerance for changes in orientation ($\alpha$), and vice versa, as shown in Fig.~\ref{fig:definition} (b).
Parameter $y$ represents the tradeoff between $\rho$ and $\alpha$.
If $y=0$, we aim to maximize the tolerance for angular deviations $\alpha$ and the tolerance for deviations in positions $\rho$ can be arbitrarily small, whereas if $y=1$ we aim to maximize $\rho$.
For $y$ in between 0 and 1, we seek values for $\alpha$ and $\rho$ within their respective ranges. 

Our start pose robustness metric allows physicians to determine how precisely they have to position the steerable needle at a planned start pose for a successful handoff to the robot. Physicians are able to adapt the tradeoff between position and orientation according to the constraints posed by a specific deployment location and according to their level of confidence between positional accuracy and orientational accuracy. 
\section{Methods}

In this section, we outline our strategy to efficiently determine a region of safe start poses $\mathbf{Q}_1$ surrounding the start pose $\mathbf{q}_1$ of a nominal motion plan $\Pi$ within start region $\Sigma$ from which the target can be reached. More specifically, our method guarantees that a collision-free plan can be found from any of the safe start poses to the target.

The steerable needle's ability to compensate for deviations from the original plan is limited by its maximum curvature constraint. 
Therefore, the needle's ability to compensate for deviations is smallest near the target but increases earlier in the trajectory.
To determine safe start poses $\mathbf{Q}_1$, we analyze the nominal motion plan starting from its final target pose. We backtrack along the plan and determine at each step how much deviation in both position and orientation would be possible such that the final pose can still be reached. In the following, we refer to this process as range propagation, as both position and orientation tolerance ranges grow (in the absence of obstacles) along the plan when moving backward toward the start pose.

In this section, we first explain in Sec.~\ref{sec:dubins} how we use 2D Dubins paths to determine poses from which the target can be reached. We describe how we use this technique to propagate ranges from one position to another in Sec.~\ref{sec:rangePropagation}. Then, in Sec.~\ref{sec:planRanges}, we outline an algorithm to efficiently propagate angular ranges from the target backward along an existing steerable needle motion plan. Finally, in Sec.~\ref{sec:surface}, we describe how to project the calculated ranges onto the start region to determine the safe start poses in $\mathbf{Q}_1$ and how we use the size of the resulting safe start region as a start pose robustness metric. 
\subsection{Dubins Paths}
\label{sec:dubins}

\begin{figure*}[t]
 \centering
 \includegraphics[width=1.0\textwidth]{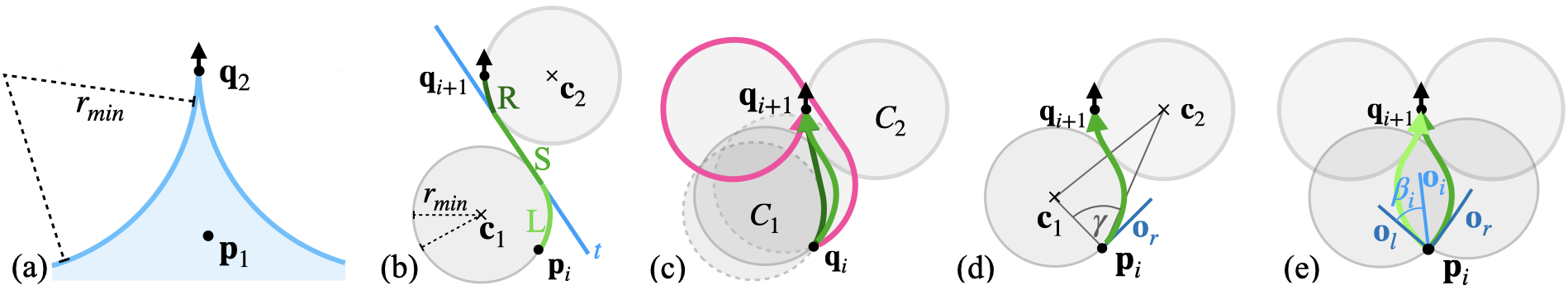}
 \caption{(a) The backward reachable workspace (blue) of $\mathbf{q}_{i+1}$ consists of all positions from which $\mathbf{q}_{i+1}$ can be reached. It is limited by arcs of radius $r_\text{min}$, resulting in a trumpet-like shape. (b) We construct an $LSR$ Dubins path connecting $\mathbf{p}_i$ and $\mathbf{q}_{i+1}$ consisting of a left curve $L$ along circle $C_1$ centered at $\mathbf{c}_1$, a straight section $S$ along inner tangent $t$ and a right curve $R$ along circle $C_2$ centered at $\mathbf{c}_2$. (c) We achieve the rightmost orientation at $\mathbf{p}_i$ if $C_1$ (solid outline) is tangent to $C_2$ (light green path). Moving $C_1$ further away reduces the orientation to the right (dark green path). Rotating the orientation further to the right results in the shortest path (pink), violating the total curvature constraint. (d) We construct a triangle $\mathbf{c}_1$, $\mathbf{c}_2$, $\mathbf{p}_i$ to find the position of $\mathbf{c}_1$ and calculate the orientation vector $\mathbf{o}_r$. (e) We repeat this process for the leftmost orientation vector $\mathbf{o}_l$ by constructing an RSL path, which results in orientation range $\phi_i$ spanning orientations between $\mathbf{o}_l$ and $\mathbf{o}_r$.}
 \label{fig:dubins}
 \end{figure*}

Our strategy for computing tolerance ranges for positions and orientations around an existing motion plan draws inspiration from Dubins Paths \cite{dubins1957curves}. In 2D, Dubins Paths are the shortest paths between any two poses $\mathbf{q}_i, \mathbf{q}_{i+1} \in \mathcal{SE}(2)$ for a system with a curvature constraint. Depending on the spatial relationship between the poses, the shortest path is one of six types of paths, which consist of straight portions (S) and left (L) and right (R) turns of maximum curvature. Shkel and Lumelsky~\cite{shkel2001classification} provide a classification system to determine which of the six path types is the shortest depending on the start and end poses' orientations relative to each other. 
Our case is not a typical application for Dubins paths, as we do not try to find the shortest path between $\mathbf{q}_i$ and $\mathbf{q}_{i+1}$. We also do not have a fixed orientation at $\mathbf{q}_i$. Instead, we are trying to determine all orientations at position $\mathbf{p}_i$ from which $\mathbf{q}_{i+1}$ can be reached.

The solution to this problem has to adhere to two steerable needle hardware constraints. The first is the needle's minimum radius of curvature constraint, which states that the needle cannot follow arcs with radii smaller than $r_\text{min}$ as its bending in tissue is limited. The second is the total curvature constraint which states that the steerable needle's total curvature between its start and final pose cannot exceed $\pi/2$ to avoid sheering effects. 
For ease of notation, we first describe our approach in 2D, i.e.~position vectors are defined as $\mathbf{p} \in \mathbb{R}^2$ and orientations are defined as $R \in \mathcal{SO}(2)$, before we explain an extension of our method to 3D in Sec.~\ref{sec:rangePropagation}.

We first analyze the positional relationship between $\mathbf{p}_i$ and $\mathbf{q}_{i+1}$. 
Due to the minimum radius of curvature constraint, $\mathbf{q}_{i+1}$ can only be reached from positions within a trumpet-like shape, as shown in Fig.~\ref{fig:dubins} (a). This area can be constructed by extending arcs with radius $r_\text{min}$ backward in the opposite direction of $\mathbf{R}_{i+1}^z$. We call the resulting area the \textit{backward reachable workspace} of $\mathbf{q}_{i+1}$. If $\mathbf{p}_{i}$ is located inside this backward reachable workspace, it means that $\mathbf{q}_{i+1}$ can be reached from $\mathbf{p}_{i}$. Therefore, we can proceed with analyzing the orientations at this position from which $\mathbf{q}_{i+1}$ is reachable.

Next, we analyze how  orientations at $\mathbf{q}_{i+1}$ influence reachability at $\mathbf{p}_i$.
We first determine the rightmost orientation at $\mathbf{p}_i$ from which $\mathbf{q}_{i+1}$ can be reached. In the 2D case, we refer to a rotation in the clockwise (negative) direction as `turning right' to align with terminology in Dubins Path literature. According to \cite{shkel2001classification}, the shortest path for such a situation is an $LSR$ path. Creating this path requires constructing two circles $C_1$ and $C_2$ with centers $\mathbf{c}_1$ and $\mathbf{c}_2$ and radius $r_\text{min}$. Circle $C_1$ intersects position $\mathbf{p}_i$ and $C_2$ is tangent to pose $\mathbf{q}_{i+1}$ (aligned in both position and orientation). Connecting the two circles by an inner tangent $t$ results in a path $C_1$-$tS$-$C_2$ connecting $\mathbf{p}_1$ and $\mathbf{p}_{i+1}$, as visualized in Fig.~\ref{fig:dubins} (b). 

 As $C_2$ is aligned with $\mathbf{q}_{i+1}$ in both position and orientation, there are only two options for its position, and the requirement of a right curve connecting to $\mathbf{q}_{i+1}$ reduces this to one fixed position. However, there are multiple possible positions for $C_1$, as only the position $\mathbf{p}_i$ is fixed, but not the orientation. Fig.~\ref{fig:dubins} (c) shows how the orientation at $\mathbf{p}_i$ changes when $c_1$ moves. The rightmost orientation at $\mathbf{p}_i$ can be achieved when both circles are tangent to each other, and the straight part of the plan is reduced to zero length. It is impossible to construct an $LSR$ plan that results in an orientation more to the right, as in that case, the circles would overlap, and no inner tangent would exist. Instead, to achieve this orientation, plan type LSL is the shortest path according to \cite{shkel2001classification}. But this path requires the needle to curve more than 360 degrees, as shown in Fig.~\ref{fig:dubins} (c), thus violating the constraint that the needle cannot curve more than $\pi/2$. As this is the shortest path connecting $\mathbf{p}_i$ to $\mathbf{q}_{i+1}$ with this orientation at $\mathbf{p}_i$, any other path connecting the two would be even longer, and also violate the total curvature constraint. Therefore, the rightmost orientation at $\mathbf{p}_i$ from which $\mathbf{q}_{i+1}$ can be reached without violating any constraints is an $LSR$ path constructed with $C_1$ and $C_2$ tangent to each other.

To find the position of $C_1$, we construct a triangle with corners $\mathbf{p}_i$, $\mathbf{c}_1$ and $\mathbf{c}_2$, as depicted in Fig.~\ref{fig:dubins} (d). Positions $\mathbf{p}_i$ and $\mathbf{c}_2$ are known, as are the side lengths $|\overrightarrow{\mathbf{p}_i\mathbf{c}_1}| = r_\text{min}$, $|\overrightarrow{\mathbf{c}_1\mathbf{c}_2}| = 2r_\text{min}$ and $|\overrightarrow{\mathbf{p}_i\mathbf{c}_2}| = ||\mathbf{c}_2 - \mathbf{p}_i||_2$ = d. Using the law of cosines we compute angle $\gamma$, the triangle's interior angle at $\mathbf{p}_i$ 
 \begin{equation*}
     \cos(\gamma) = \frac{-3r_\text{min}^2 + d^2}{2r_\text{min}d}.
 \end{equation*}
 This allows us to determine position $\mathbf{c}_1$. Finally, we calculate
 orientation $\mathbf{o}_r \in \mathcal{SO}(2)$,
 the start orientation of the $LSR$ path at $\mathbf{p}_i$, which is perpendicular to $\overrightarrow{\mathbf{c}_1\mathbf{p}_i}$. Similarly, we connect $\mathbf{p}_i$ and $\mathbf{q}_{i+1}$ with an RSL path to determine $\mathbf{o}_{l}$, the maximum orientation to the left at $\mathbf{p}_i$, as shown in Fig.~\ref{fig:dubins} (e). 
 
 We define an \textit{orientation range} as a tuple 
 $\phi_i = \{\mathbf{o}_i, \beta_i\}$, 
 where $\mathbf{o}_i \in \mathcal{SO}(2)$ is an orientation vector, and $\beta \in [-\pi/2, \pi/2]$ is an angle representing the maximum deviation from $\mathbf{o}_i$. 
 We define the orientation range at $\mathbf{p}_i$ to represent the boundary orientations we computed: center orientation $\mathbf{o}_i = (\mathbf{o}_l + \mathbf{o}_r) / ||  \mathbf{o}_l + \mathbf{o}_r ||_2$ 
 and angle $\beta = \cos^{-1} (\mathbf{o}_i \cdot \mathbf{o}_l)$. As shown in Fig.~\ref{fig:dubins} (e), for any orientation within $\phi_i$, a Dubins path from $\mathbf{p}_i$ to $\mathbf{q}_{i+1}$ adhering to all curvature constraints exists and $\mathbf{q}_{i+1}$ is therefore reachable.
\subsection{Range Propagation}
\label{sec:rangePropagation}
In Sec.~\ref{sec:dubins}, we determined the orientation range $\phi_i$ at $\mathbf{p}_i$ by considering a single orientation at $\mathbf{q}_{i+1}$. Now, we extend the orientation range calculation to consider an orientation range $\phi_{i+1}$ at $\mathbf{p}_{i+1}$.
Our goal is to find the maximum orientation range at $\mathbf{p}_i$ from which $\mathbf{p}_{i+1}$ can be reached with an orientation within $\phi_{i+1}$.
We refer to this process as \textit{propagating orientation ranges} from $\phi_{i+1}$ to $\mathbf{p}_i$.
An orientation range $\phi_{i+1}$ at a position $\mathbf{p}_{i+1}$ is \textit{reachable} from a position $\mathbf{p}_i$ if at least one path adhering to all constraints can be found from $\mathbf{p}_i$ to $\mathbf{p}_{i+1}$ whose orientation at $\mathbf{p}_{i+1}$ is within the orientation range $\phi_{i+1}$. It is important to note that not all orientations within $\phi_{i+1}$ have to be reached, a subset suffices.
 We define function
 \begin{equation}
\text{RangeDistance}(\mathbf{o}, \phi_i) = \begin{cases}
0  &  \text{if} \ |\delta| < \beta_i \\
\delta + \beta_i &  \text{if} \ \delta < -\beta_i \\
\delta - \beta_i &  \text{if} \ \delta > +\beta_i
\end{cases},
\label{eq:inRange}
\end{equation}
where $\delta \in [-\pi/2, \pi/2]$ is the signed angle between orientation vector $\mathbf{o}$ and the center orientation vector of $\phi_i$, $\mathbf{o}_i$, such that $Rot(\delta) \mathbf{o}_i = \mathbf{o}$, with $Rot(\delta) \in SO(2)$ being a 2D rotation about angle $\delta$ and $\beta_i$ is an angle representing the deviation from $\mathbf{o}_i$, as defined in Sec.~\ref{sec:dubins}. The function returns $0$ if $\mathbf{o}$ is within the orientation range. A positive range distance corresponds to an orientation beyond the left border of the orientation range, whereas a negative range distance is caused by an orientation to the right of the orientation range's right border. 

\begin{figure}[t]
 \centering
 \includegraphics[width=1.0\columnwidth]{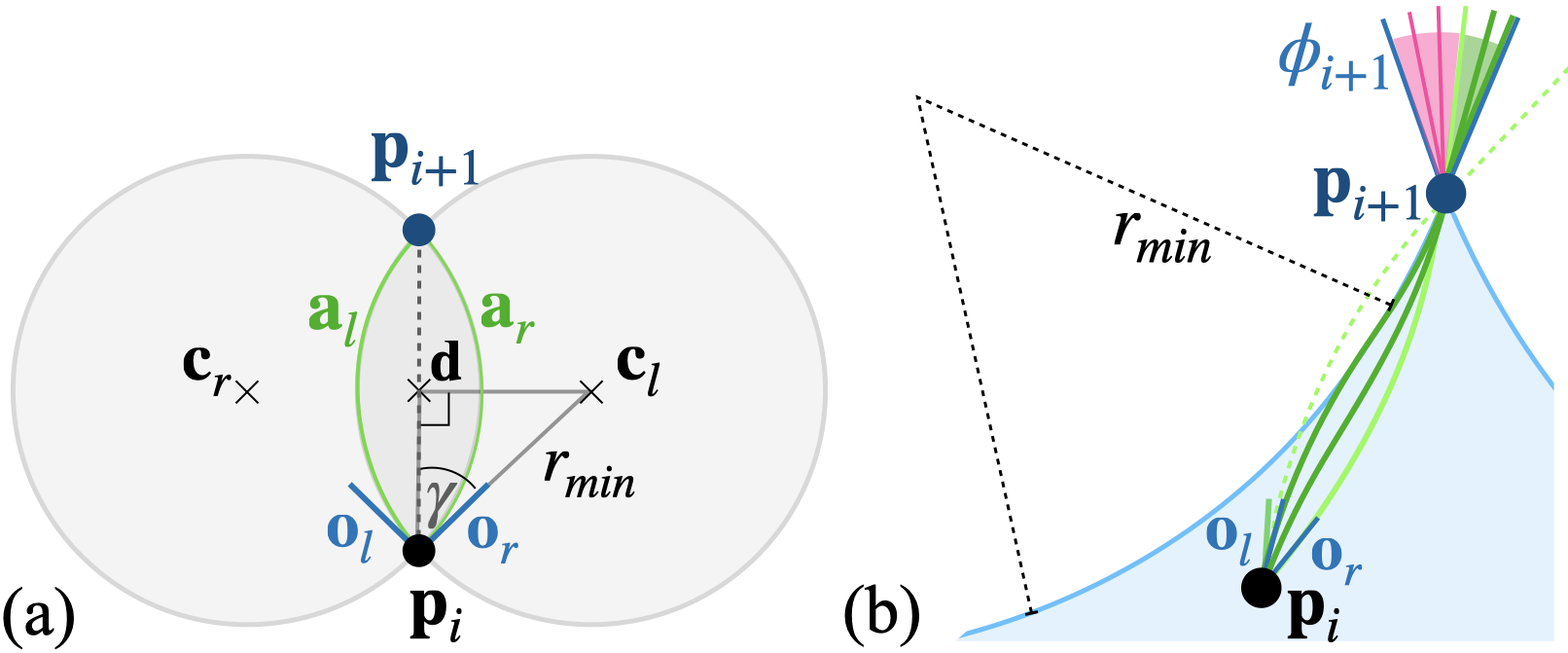}
 \caption{(a) We construct two constant curvature arcs $\mathbf{a}_l$ and $\mathbf{a}_r$ with radius of curvature $r_\text{min}$ connecting $\mathbf{p}_i$ and $\mathbf{p}_{i+1}$, resulting in orientations $\mathbf{o}_l$ and $\mathbf{o}_r$. This is the maximum orientation range that could be propagated from $\mathbf{p}_{i+1}$ without taking orientation range $\phi_{i+1}$ into account. (b) Orientation range $\phi_{i+1}$ (dark blue) can be reached from positions $\mathbf{p}_i$ within its backward reachable workspace (light blue) limited by arcs of radius $r_\text{min}$. We connect position $\mathbf{p}_i$ to $\mathbf{p}_{i+1}$ with a pair of constant curvature arcs (light green) to propagate the orientation range $\phi_{i+1}$ to $\mathbf{p}_i$. As arc $\mathbf{a}_l$ (dashed lines) lies partially outside the backward reachable workspace, we replace them with an RSL Dubins path (dark green). Some orientations in $\phi_{i+1}$ (pink) cannot be reached from $\mathbf{p}_i$.
 }
 \label{fig:rangePropagation}
 \end{figure}

Similar to the single orientation case, we determine the backward reachable workspace of $\phi_{i+1}$ by extending two constant curvature arcs with radius $r_\text{min}$ starting at $p_{i+1}$ and their orientations aligned with $\mathbf{o}_{i+1,l}$  and $\mathbf{o}_{i+1,r}$, respectively. Positions outside of this reachable workspace cannot reach $\mathbf{o}_{i+1}$ with an orientation within orientation range $\phi_{i+1}$. 
If $\mathbf{p}_i$ is located inside the backward reachable workspace, we first attempt to connect $\mathbf{p}_{i+1}$ and $\mathbf{p}_i$ with a constant curvature arc of radius $r_\text{min}$. There are two possible solutions for connecting two points by an arc with a known radius, and we refer to these as $\mathbf{a}_l$ and $\mathbf{a}_r$.
Next, we describe the process for constructing $\mathbf{a}_l$ as shown in Fig.~\ref{fig:rangePropagation} (a); $\mathbf{a}_r$ can be found by mirroring this process. We find the center of $\mathbf{a}_l$ by constructing a triangle with corners $\mathbf{p}_i$, $\mathbf{c}_l$ and $\mathbf{d}$, where $\mathbf{c}_l$ is the center of the circle the arc follows and $\mathbf{d}$ is the midpoint between $\mathbf{p}_i$ and $\mathbf{p}_{i+1}$. The triangle has two known lengths $|\overrightarrow{\mathbf{p}_i\mathbf{c}_l}| = r_\text{min}$ and $|\overrightarrow{\mathbf{p}_i\mathbf{d}}| = ||\mathbf{p}_{i+1} - \mathbf{p}_i||_2/2 = d/2$ and a $90^\circ$ interior angle at $\mathbf{d}$. Therefore, we can solve for the interior angle at $\mathbf{p}_i$ as $\gamma = \cos^{-1} (\frac{d/2}{ r_\text{min}})$. Now, we can calculate orientation $\mathbf{o}_l$,
 the start orientation of the arc at $\mathbf{p}_i$, which is perpendicular to $\overrightarrow{\mathbf{c}_l\mathbf{p}_i}$. Arc $\mathbf{a}_l$ can be interpreted as the extreme case of an RSL Dubins path (see Sec.~\ref{sec:dubins}) with straight and left-curving sections of length zero. Therefore, the resulting orientation $\mathbf{o}_l$ is the leftmost possible orientation at $\mathbf{p}_i$ from which $\mathbf{p}_{i+1}$ can be reached regardless of the orientation at $\mathbf{p}_{i+1}$.
 For an orientation further to the left, a connecting path would either have to have a smaller radius of curvature, violating the minimum radius of curvature constraint, or perform a $360^{\circ}$ loop (see Fig.~\ref{fig:dubins}), violating the total curvature constraint. 

So far, we have only considered the position of $\mathbf{p}_{i+1}$, but not its orientation range $\phi_{i+1}$, which introduces additional restrictions, as shown in Fig.~\ref{fig:rangePropagation} (b).
We check if the arcs' orientation at $\mathbf{p}_{i+1}$ is within orientation range $\phi_{i+1}$. 
To test $\mathbf{a}_l$, we compute $\text{RangeDistance}(\mathbf{o}_{i+1}, \phi_{i+1}) = \epsilon$ (see Eq.~\ref{eq:inRange}), where $\mathbf{o}_{i+1} \in \mathcal{SO}(2)$ is the arc's orientation at $\mathbf{p}_{i+1}$. 
If $\epsilon=0$, $\mathbf{o}_{i+1}$ is within orientation range $\phi_{i+1}$. This is equivalent to arc $\mathbf{a}_l$ lying fully inside $\mathbf{p}_{i+1}$'s backward reachable workspace; therefore, it connects $\mathbf{p}_i$ to $\mathbf{p}_{i+1}$ and adheres to all curvature constraints.

However, if $\epsilon < 0$, $\mathbf{a}_l$ curves too far to the right and is not in range $\phi_{i+1}$. In this case, we replace the arc with a Dubins Path as explained in Sec.~\ref{sec:dubins}. We seek to find the leftmost possible orientation at $\mathbf{p}_i$ and therefore do not want to rotate $\mathbf{o}_l$ more to the right than necessary. Arc $\mathbf{a}_l$ curving to the right is the most extreme case of an RSL path with no S or L sections (see Sec.~\ref{sec:dubins}). Therefore, we replace $\mathbf{a}_l$ with an RSL path connecting to the rightmost orientation of $\phi_i$, as can be seen in Fig.~\ref{fig:rangePropagation} (b). We rotate $\mathbf{o}_l$ to the right to align with the RSL path's start orientation, thus reducing the orientation range at $\mathbf{p}_i$.
We define function 
 \begin{equation}
 \text{RSL}(\mathbf{p}_i, \mathbf{p}_{i+1}, \phi_{i+1}) = \mathbf{o}_l,
 \label{eq:RSL}
 \end{equation} 
 which computes $\mathbf{o}_{l}$, the leftmost orientations at $\mathbf{p}_i$, from which $\mathbf{p}_{i+1}$ can be reached within orientation range $\phi_{i+1}$. This function first attempts to find a constant curvature arc connection and replaces it with another RSL path if necessary, as described above.
 
Similarly, we define 
\begin{equation}
 \text{LSR}(\mathbf{p}_i, \mathbf{p}_{i+1}, \phi_{i+1}) = \mathbf{o}_r,
 \label{eq:LSR}
 \end{equation}
 which determines the rightmost orientation $\mathbf{o}_r$ at $\mathbf{p}_i$, from which $\mathbf{p}_{i+1}$ can be reached within orientation range $\phi_{i+1}$. This function mirrors the process described above. We first construct arc $\mathbf{a}_r$ and check if it connects to $\mathbf{p}_{i+1}$ with an orientation within orientation range $\phi_{i+1}$. If  $\text{RangeDistance}(\mathbf{o}_{i+1}, \phi_{i+1}) > 0$, $\mathbf{a}_r$ curves too far to the left. In this case, we replace it with a Dubins LSR path and update $\mathbf{o}_r$ accordingly.
An edge case occurs if $\mathbf{p}_i$ is located on the boundary of the backward reachable workspace. In this case, there is only one possible path connecting to $\mathbf{p}_{i+1}$ within orientation range $\phi_{i+1}$, which is an arc following the workspace boundary. In this case, there is only a single orientation at $\mathbf{p}_i$ and $\mathbf{o}_l$ and $\mathbf{o}_r$ are identical with no tolerance for deviation. 

Once we have found start orientations $\mathbf{o}_l$ and $\mathbf{o}_r$ at $\mathbf{p}_i$ that connect to $\phi_{i+1}$, we use them to define orientation range $\phi_i = \{\mathbf{o}_{i},\beta\}$. From any orientation within $\phi_i$, a connection to at least one orientation in $\phi_{i+1}$ at $\mathbf{p}_{i+1}$ exists.
 \subsection{Propagating Orientation Ranges along a Motion Plan}
\label{sec:planRanges}

\begin{figure*}[t]
 \centering
 \includegraphics[width=0.7\textwidth]{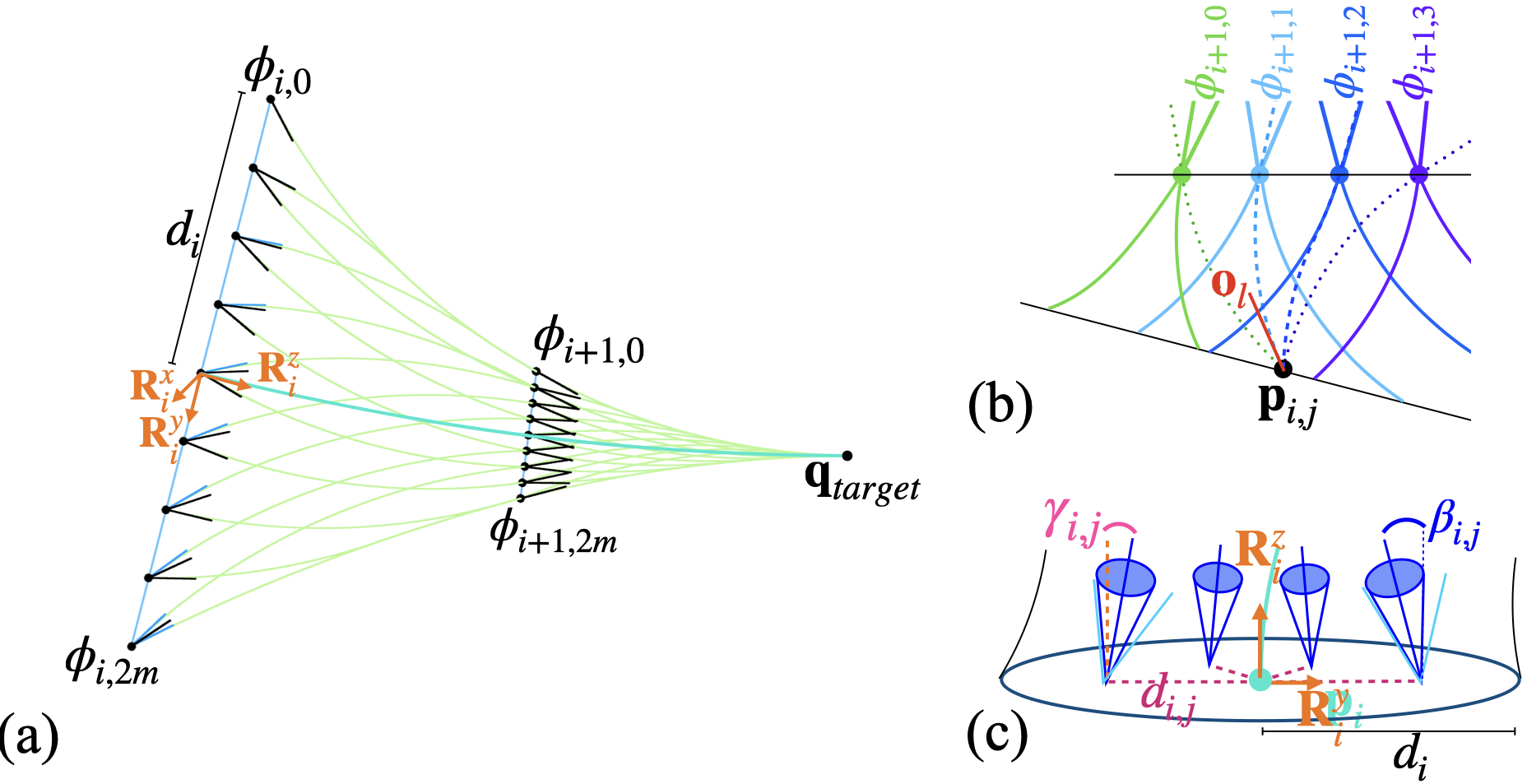} 
 \caption{(a) Propagating orientation ranges backward along the plan starting at $\mathbf{q}_\text{target}$. We propagate orientation ranges $\Phi_{i+1}$ to create orientation ranges $\Phi_{i}$. (b) We try to find orientation ranges as wide as possible. To determine the left boundary $\mathbf{o}_l$ at $\mathbf{p}_i$, we find the leftmost $\phi_{i+1}$ to which an RSL connection (dotted lines) within their respective backward reachable workspaces (solid lines) exists. Here, $\phi_{i+1,0}$ is out of range, $\phi_{i+1,1}$ results in the leftmost $\mathbf{o}_l$, a connection less to the left exists to $\phi_{i+1,2}$, and $\phi_{i+1,3}$ is again unreachable. c) Orientation ranges in 3D (dark blue) are defined by their distance $d_{i,j}$ to the nominal position $\mathbf{p}_i$ (cyan), by the angle $\gamma_{i,j}$ between their center orientation and the deployment direction $\mathbf{R}_i^z$, and
 by the spread of the angular deviation $\beta_{i,j}$. After determining the initial boundaries of orientation ranges in 2D (light blue), we shrink the ranges to achieve symmetry around $\mathbf{R}_i^z$.}
 \label{fig:planRanges}
 \end{figure*}

In this section, we introduce an algorithm that efficiently propagates orientation ranges along an existing motion plan $\Pi$. We begin from $\mathbf{q}_\text{target}$ and we propagate orientation ranges along $\Pi$ until we reach the start pose $\mathbf{q}_1$. For each plan section $\mathbf{a}_i$, our algorithm calculates an ordered list of $2m+1$ orientation ranges $\Phi_i = [\phi_{i,0}, \ \dots, \phi_{i,j}, \dots, \phi_{i,2m}], \ j,m \in \mathbb{N}$ associated with plan configuration $\mathbf{q}_i$.

The nominal motion plan $\Pi$ was created in a 3D environment. However, each plan section $\mathbf{a}_i$ consists of a constant curvature arc in a 2D plane, and we perform our range propagation strategy in this plane spanned by the needle's deployment direction $\mathbf{R}_i^z$ and $-\mathbf{R}_i^y$, the direction it is curving into (see Fig.~\ref{fig:planRanges} (a)). Our strategy computes conservative orientation ranges based on their distance to $\mathbf{p}_i$ such that they can be extended into 3D, as explained later in this section. This allows us to apply our method to each plan section, as their constant curvature arcs curve in different 2D planes.

Beginning at the target pose, we define a single orientation range $\Phi_\text{target} = [\{ \mathbf{o}_\text{target}, 0 \}]$ with a single orientation vector $\mathbf{o}_\text{target}$, which corresponds to the final orientation of the nominal plan $\mathbf{q}_\text{target}$. We set the angular deviation to zero as this is the orientation we aim to converge towards when the steerable needle is deployed.

For each previous plan section, we first determine the backward reachable workspace of $\phi_{i+1,0}$ at $\mathbf{p}_{i+1,0}$, which is the leftmost orientation range, the direction $-\mathbf{R}_i^y$ in which the current plan section $\mathbf{a}_i$ is curving. Positions outside the left limit of $\phi_{i+1,0}$'s backward reachable workspace cannot be reached from any orientation range in $\Phi_{i+1}$ as it would require a smaller radius of curvature than $r_\text{min}$. 

As our goal is to ensure a safe needle insertion without any collisions, we have to check for obstacles in the proximity of the start pose. We take steps of step size $s$ along plan section $\mathbf{a}_i$ starting at $\mathbf{p}_{i+1}$, and for each position $\mathbf{p}_s \in \mathbb{R}^3$, we determine the minimum distance $d_s$ to the left boundary of the backward reachable workspace. We compute the minimum distance of $\mathbf{p}_s$ to an obstacle as \begin{equation*}
    f_s = \underset{v \in \mathcal{O}}{\mathrm{argmin}} \ || \mathbf{p}_s - \mathbf{v}||_2 -r_v - \frac{d_\text{needle}}{2},
\end{equation*}
where $\mathbf{v} \in \mathbb{R}^3$ is the center position of a sphere containing obstacle $v \in \mathcal{O}$, $r_v \in \mathbb{R}$ is the obstacle sphere's radius, $d_\text{needle}$ is the steerable needle's diameter, and $f_s \in \mathbb{R}$ is the resulting minimum obstacle distance. This is a conservative measure that ensures collision avoidance. If $f_s$ is smaller than $d_s$, an obstacle is located inside the backward reachable workspace. In this case, we shrink $d_s$ to $f_s$, and we split $\mathbf{a}_i$ into two sections $\mathbf{a}_i$ and $\mathbf{a}_i'$, the second of which starts at $\mathbf{p}_s$. More details about this process can be found in \cite{hoelscher2022metric}. 

 In the following section, we outline our algorithm for efficient range propagation from $\Phi_{i+1}$ to $\Phi_i$ along one plan section $\mathbf{a}_i$ in Alg.~\ref{alg:range_propagation}.
For each previous plan section, we define the positions of orientation ranges $\Phi_i$ (Alg.~\ref{alg:range_propagation}, line \ref{alg:positions}), which we refer to as $\mathbf{P}_i \in \mathbb{R}^3$.
We determine the minimum distance $d_i$ between the left workspace boundary of $\phi_{i+1}$ and $\mathbf{q}_i$ in negative y-direction $-\mathbf{o}^y_i$, the direction the constant curvature arc is curving into. 
We evenly split $d_i$ into $m$ steps of step size $d_i/m$ each, and we position a total of $2m+1$ orientation ranges in both positive and negative y-direction from $\mathbf{q}_i$, as shown in Fig.~\ref{fig:planRanges} (a).

Now, we propagate the orientation ranges from $\Phi_{i+1}$ to create orientation ranges $\Phi_{i}$. Unlike when we introduced the concept of orientation propagation in Sec.~\ref{sec:rangePropagation}, we now face a situation with $2m+1$ orientation ranges in $\Phi_{i+1}$ to be propagated to create $2m+1$ new ranges associated with $\mathbf{q}_i$.
It would be computationally expensive to propagate every $\phi_{i+1,j} \in \Phi_{i+1}$ to every $\phi_{i,k} \in \Phi_i$ to determine the maximum orientation range for each. Therefore, we introduce an efficient algorithm that requires significantly fewer orientation propagations. 

We explain the process for propagating the left limits of orientation ranges, as shown in Fig.~\ref{fig:planRanges} (b). We can find the leftmost orientation $\mathbf{o}_l$ at $\mathbf{p}_i$ by connecting it to $\phi_{i+1}$ as far to the left as possible. Therefore, we begin our search at $\mathbf{p}_{i,0}$ and $\phi_{i+1,0}$ and work from the left to the right (Alg.~\ref{alg:range_propagation}, line \ref{alg:loopI}-\ref{alg:loopI1}).
 
As described in Sec.~\ref{sec:rangePropagation}, we begin by connecting $\mathbf{p}_{i,j}$ and $\mathbf{p}_{i+1,k}$ with a right-curving constant curvature arc of radius $r_\text{min}$ (Alg.~\ref{alg:range_propagation}, line \ref{alg:arcRight}), resulting in orientation $\mathbf{o}_{i+1}$ at $\mathbf{p}_{i+1,k}$. Now, we compute the range distance between $\mathbf{o}_{i+1}$ and $\phi_{i+1,k}$ as defined in Eq.~\ref{eq:inRange} (Alg.~\ref{alg:range_propagation}, line \ref{alg:rangeDistanceLeft}). If the range distance is larger than zero, $\mathbf{o}_{i+1}$ is pointing too far to the left with respect to $\phi_{i+1,k}$ and no connection with $\mathbf{p}_{i,j}$ can be found, as visualized in Fig.~\ref{fig:rangePropagation} (b) in pink. Since we started creating orientation ranges at positions $\mathbf{p}_i$ from the left, any subsequent $\mathbf{p}_{i,j+1}$ will be located further to the right. Therefore a connection between $\mathbf{p}_{i,j+1}$ and $\mathbf{p}_{i+1,k}$ would also be out of range $\phi_{i+1,k}$ and we can skip testing connections to $\mathbf{p}_{i+1,k}$ for all subsequent $\mathbf{p}_{i,j+1}$ (Alg.~\ref{alg:range_propagation}, line \ref{alg:skip}). 

If the range is not larger than zero, we next check if $\mathbf{p}_{i,j}$ lies within the backward reachable workspace of $\phi_{i+1,k}$ (Alg.~\ref{alg:range_propagation}, line \ref{alg:workspace}). If this is the case, a path between $\mathbf{p}_{i,j}$ and $\phi_{i+1,k}$ exists. Therefore, we apply the RSL function as defined in Eq.~\ref{eq:RSL}, 
 (Alg.~\ref{alg:range_propagation}, line \ref{alg:RSL}). Since we started testing orientation ranges $\phi_{i+1,k}$ from the left, the first RSL path found results in the leftmost possible orientation $\mathbf{o}_l$ at $\mathbf{p}_{i,j}$ as can be seen in the example in Fig.~\ref{fig:planRanges} (b) in light blue. Connecting $\mathbf{p}_{i,j}$ to any other orientation range further to the right would result in an orientation pointing less to the left (see Fig.~\ref{fig:planRanges} (b) in dark blue). In case the step size between positions is too large, it can happen that no connection between $\mathbf{p}_{i,j}$ and an orientation range $\phi_{i+1,k}$ can be found, in which case we leave the orientation range at $\mathbf{p}_{i,j}$ empty.
 We repeat this process for all subsequent $\phi_{i,j}$ until we reach the rightmost range.
The process for right boundaries is very similar, but starting at the rightmost position $\mathbf{p}_{i,2m}$ and iterating over angular ranges from right to left (Alg.~\ref{alg:range_propagation}, line \ref{alg:loop2}) finding LSR paths (Alg.~\ref{alg:range_propagation}, line \ref{alg:LSR}). After we found both left and right boundaries at each $\mathbf{p}_{i,j}$, we combine them to orientation ranges $\Phi_{i,j}$ (Alg.~\ref{alg:range_propagation}, line \ref{alg:range}).

Although we analyze plan sections in the 2D plane they are curving in, this plane is potentially changing with each plan section. Therefore, our method has to be generalizable into 3D. Our previous orientation range definition $\phi_i = \{\mathbf{o}_{i},\beta\}$ in 3D results in a cone shape with center orientation $\mathbf{o}_{i}$ and an angular deviation $\beta$ from $\mathbf{o}_{i}$ in any direction. As explained above, the positions of orientation ranges $\Phi_i$ are limited by the backward reachable workspace boundary in the direction of plan curvature $-R_y^i$ by the minimum radius of curvature $r_\text{min}$. To expand into 3D, we determine the distance $d_i$ between $\Phi_i$ and nominal plan position $p_i$, as shown in Fig.~\ref{fig:planRanges} (c). Then, we form a circle of radius $d_i$ in the plane perpendicular to deployment direction $R_z^i$.

Now, we define orientation range positions by their distance $d_{i,j}$ to $\mathbf{p}_i$, no matter where on the circle they are located, to achieve symmetry around the nominal motion plan. We express the center orientation $\mathbf{o}_{i}$ by signed angle $\gamma$, the angle between $\mathbf{o}_{i}$ and deployment orientation $R_i^z$, with positive values of $\gamma$ expressing orientations pointing away from the center and negative values oriented towards the center. 

In order to create orientation ranges symmetric around $\mathbf{R}_i^z$ we have to shrink the orientation ranges previously computed in 2D. The orientation ranges in the direction of the plan curvature and in the opposite direction are the most extreme cases for possible orientations at a distance $d_{i,j}$ from the center. Therefore, enforcing symmetry between them results in orientation ranges that can be positioned anywhere on the circle spanned by distance $d_{i,j}$, and it will be possible to find a path from these ranges to the orientation ranges of the next plan section, as depicted in Fig.~\ref{fig:planRanges} (c). This centrosymmetric approach creates conservative orientation ranges in 3D around the nominal motion plan backward from $\mathbf{q}_\text{target}$ to $\mathbf{q}_1$.

 \restylefloat{algorithm} 
\begin{algorithm}[t]
    \small
\caption{Range Propagation}
 \label{alg:range_propagation}
\begin{algorithmic}[1]
 \Function{RangePropagation}{$\mathbf{q}_i, \ \mathbf{a}_i, \ \Phi_{i+1}$}

 \State $\mathbf{P}_{i} \leftarrow  \text{GetPositions}(\mathbf{q}_i, \ \Phi_{i+1})$
 \label{alg:positions}
 \State $c$ $\leftarrow$ 0
 \For{$j=0:2m$} \Comment{Left limits} \label{alg:loopI}
    \For{$k=c:2m$} \label{alg:loopI1}
        \State $\mathbf{o}_{i+1} = \text{ArcConnectRight} (\mathbf{p}_{i,j}, \ \mathbf{p}_{i+1,k})$ \label{alg:arcRight}
        \If {$\text{RangeDistance}(\mathbf{o}_{i+1}, \phi_{i+1,k}) > 0$} \label{alg:rangeDistanceLeft}
             
            \State $c\leftarrow c+1$ \label{alg:skip}
        
        \Else
            \If {$\text{InBackWorkspace}(\mathbf{p}_{i,j}, \ \mathbf{p}_{i+1,k}, \ \phi_{i+1,k})$} \label{alg:workspace}
            \State $\mathbf{o}_{j,l} \leftarrow  \text{RSL}(\mathbf{p}_{i,j}, \ \mathbf{p}_{i+1,k}, \ \phi_{i+1,k})$ \label{alg:RSL}
            \EndIf
             \State \textbf{break}      
        \EndIf     
    \EndFor
\EndFor

\For{$j=2m:-1:0$} \Comment{Right limits}
    \For{$k=c:-1:0$} \label{alg:loop2}
        \State $\mathbf{o}_{i+1} = \text{ArcConnectLeft}(\mathbf{p}_{i,j}, \ \mathbf{p}_{i+1,k})$ \label{alg:arcLeft}
        \If {$\text{RangeDistance}(\mathbf{o}_{i+1}, \phi_{i+1,k}) < 0$} \label{alg:rangeDistanceRight}
            \State $c \leftarrow c-1$ 
            \label{alg:skipRight}
        \Else 
        \label{alg:workspaceRight}
            \If {$\text{InBackWorkspace}(\mathbf{p}_{i,j}, \ \mathbf{p}_{i+1,k}, \ \phi_{i+1,k})$}
                \State $\mathbf{o}_{j,r} \leftarrow \text{LSR}(\mathbf{p}_{i,j}, \ \mathbf{p}_{i+1,k}, \ \phi_{i+1,k})$ 
                \label{alg:LSR}
            \EndIf
             \State \textbf{break}
        
        \EndIf 
    \EndFor
\EndFor
\For{$j=0:2m$}
    \State $\phi_{i,j} \leftarrow \text{OrientationRange}(\mathbf{o}_{j,l}, \ \mathbf{o}_{j,r})$ \label{alg:range}
\EndFor
 \State \Return $\Phi_{i}$ 
\EndFunction
 \end{algorithmic}
 \end{algorithm}

\subsection{Start Region Evaluation and Robustness Metric}
\label{sec:surface}

Using the algorithm outlined in Sec.~\ref{sec:planRanges}, we computed orientation ranges for each plan section in nominal motion plan $\Pi$, including the first plan section, resulting in orientation ranges $\Phi_1$ that represent alternative start poses around nominal start pose $\mathbf{q}_1$ from which a collision-free path to the target can be found. In this subsection, we explain how to determine the intersection of sets $\Sigma$, the start poses that a physician can reach, and $\Phi_1$. The result of this process is $\mathbf{Q}_1$, the region of safe start poses from which a collision-free path to the target can be found and which can be reached by a physician. We also define the robustness metric $R$ to evaluate $\Pi$ based on the size of $\mathbf{Q}_1$ in both position and orientation.

\begin{figure}[t]
 \centering
 \includegraphics[width=0.7\columnwidth]{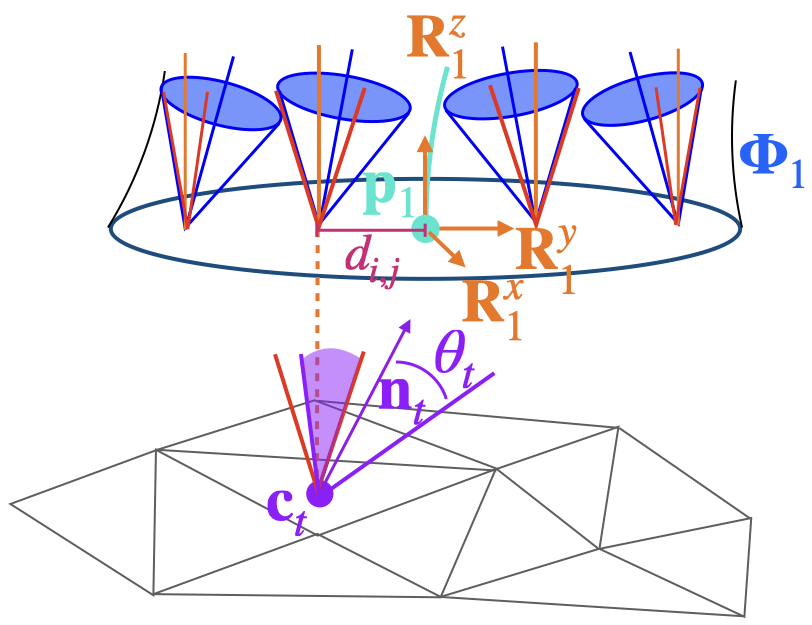}
 \caption{Assuming the physician will aim for the start orientation of the nominal motion plan $\mathbf{R}_1^z$, we shrink all orientation ranges in $\Phi_1$ (blue) to be centered around $\mathbf{R}_1^z$ (red lines).
 We represent start poses $\Sigma$ as a triangle mesh with angular ranges positioned at the centroid $\mathbf{c}_t$ of the triangles and centered around triangle normals $\mathbf{n}_t$ with angular deviation $\theta_t$. We project centroids $\mathbf{c}_t$ of triangles in the mesh onto the plane perpendicular to $\mathbf{R}_1^z$. We determine the angular range closest to the projected center, and we determine the intersection between this orientation range and the triangle's orientation range (shaded in purple). This orientation range can be reached by the physician, and a collision-free motion plan to the target can be found from it.}
 \label{fig:surface_projection}
 \end{figure}

Based on the assumption that physicians will aim to insert the steerable needle following the nominal start orientation $\mathbf{R}_1^z$, we shrink all orientation ranges in $\Phi_1$ to be centered around orientation $\mathbf{R}_1^z$, as shown in Fig.~\ref{fig:surface_projection}. This allows us to analyze the robustness of deviation from the nominal start orientation.

We represent start region poses $\Sigma$ as orientation ranges on a triangular mesh. The orientation ranges are positioned at the centroids $\mathbf{c}_t$ of the triangles $t$ and are centered around triangle normals $\mathbf{n}_t$ with angular deviation $\theta_t$ (see Fig.~\ref{fig:surface_projection}). These orientation ranges represent the poses that a physician can reach.
To find intersections between $\Sigma$ and $\Phi_1$, we use a breadth-first approach: we start analyzing the triangle $\mathbf{p}_1$ is located in, and we continue with its direct neighbors, adding them to a priority queue. As we aim to find a safe start region surrounding the nominal plan without any holes, we only expand to the neighbors of triangles whose orientation ranges are at least partially part of $\mathbf{Q}_1$ as determined in a process explained below. Therefore, the search terminates once it reaches the boundaries of the start region that is potentially part of the safe start region.

While the nominal start position $\mathbf{p}_1$ is always located in the start region by definition, this is not necessarily true for other poses in $\Phi_1$, as the start region does not have to be on a flat surface.
Therefore, we have to project the orientation ranges in $\Phi_1$ onto the start region. For each triangle we are analyzing, we project its centroid position $\mathbf{c}_t$ along initial deployment direction $\mathbf{R}_1^z$ to position $\mathbf{c}_t'$ on the plane spanned by $\mathbf{R}_1^x$ and $\mathbf{R}_1^y$ around $\mathbf{p}_1$. We calculate the distance $d$ between $\mathbf{c}_t'$ and $\mathbf{p}_1$. If this distance is smaller than the distance of the outermost orientation range to the center $d_{1,0}$, we find the orientation range closest to distance $d$, and we project it back onto the start region at $\mathbf{c}_t$. Now, we determine the overlap between the projected orientation range and the orientation range at $\mathbf{c}_t$, as shown in Fig.~\ref{fig:surface_projection}. If the intersection of the two orientation ranges is not an empty set, we add the resulting poses to $\mathbf{Q}_1$, and we add the unseen neighbors of this triangle $t$ to the priority queue. With a slight abuse of notation, we write $t \in \mathbf{Q}_1$ if the orientation range located at $\mathbf{c}_t$ contributed to poses in $\mathbf{Q}_1$.

Once we have analyzed all surrounding triangles and the breadth-first search has terminated, we aim to express the robustness of plan $\Pi$ to deviations in the start position and orientation by quantifying the region of safe start poses $\mathbf{Q}_1$. To do so, we define the start pose robustness metric
$R(\Pi, \Sigma, y)\mapsto (\rho, \alpha)$, $y \in [0,1]$, $\rho \in \mathbb{R}^+$, $\alpha \in [0,\pi]$. 
We first determine the maximum position deviation in $\mathbf{Q}_1$ as the maximum radius around $\mathbf{p}_1$ within which no triangles are located, whose orientation ranges are not part of $\mathbf{Q}_1$:
\begin{align*}
    \rho_\text{max}=\text {max } \rho \ 
    &\text{s. t.} \ ||\mathbf{c}'_t - \mathbf{p}_1||_2  > \rho \ \forall t_j \in \Sigma\setminus \mathbf{Q}_1, j \in \mathbb{N},
\end{align*}
where $\rho_\text{max} \in \mathbb{R}^+$ is the maximum positional deviation from $\mathbf{p}_1$, and $\mathbf{c}'_t$ is the projection of the centroid of triangle $t_j$ into the $\mathbf{R}_1^x$, $\mathbf{R}_1^y$ plane. 
As explained in Sec.~\ref{sec:problem}, it is impossible to maximize both the position robustness and the orientation robustness in $\mathbf{Q}_1$. Therefore, we define parameter $y$ as a user-selected trade-off between allowable position and orientation deviation. If we choose $y=0$, we maximize the robustness to changes in position, so robustness metric $R(\Pi, \Sigma, y)$ results in $\rho = \rho_\text{max}$, whereas the robustness to angular deviations $\alpha = 0$, as depicted in Fig.~\ref{fig:definition}. Tradeoff parameter $y$ is anti-proportional to $\rho$, such that $\rho = (1-y)\rho_\text{max}$. For any value $y>0$, we determine the deviation to changes in orientation as the minimum deviation angle of any orientation range in $\mathbf{Q}_1$ within radius $\rho$:
\begin{align*}
    \alpha =\text {min} \ \beta_t \ 
    &\text{s. t.} \ ||\mathbf{c}'_t - \mathbf{p}_1||_2 < \rho \ \forall t_j \in \mathbf{Q}_1, j \in \mathbb{N},
\end{align*}
where $\mathbf{c}'_t$ is again the projection of the centroid of triangle $t_j$ into the $\mathbf{R}_1^x$, $\mathbf{R}_1^y$ plane, and $\beta_t$ is the angle of the orientation range in $\mathbf{Q}_1$ located at $\mathbf{c}_t$. 
This procedure allows us to make statements such as ``If we can achieve a start pose within a radius $\rho$ from $\mathbf{p}_1$ and an orientation within $\alpha$ degrees from orientation $\mathbf{R}_1^z$, we can guarantee that a collision-free motion plan to the target can be found.'' Trade-off parameter $y$ allows users to adjust the sensitivity to deviations in positions and orientations according to the specific deployment scenario.

\section{Evaluation}
We tested our start pose robustness metric in three different motion planning scenarios for steerable needles to highlight its versatility, including an abstract scenario with spherical obstacles (Sec.~\ref{sec:spheres}), a percutaneous liver biopsy scenario (Sec.~\ref{sec:liver}), and a transoral lung access scenario (Sec.~\ref{sec:lung}).
In each of these scenarios, we identified an insertion surface consisting of potential start poses for the steerable needle, as well as critical obstacles the needle has to avoid. 
For collision detection, we 
 used a nearest neighbor search structure \cite{ichnowski2018concurrent} containing 
 the center positions of all obstacle voxels as it allows for efficient detection of close-by obstacles. If a position's distance to its nearest nearest neighbor is larger than the radius of a sphere enclosing a voxel, we consider that position as collision-free. We use this collision detection strategy for both the motion planners we used and for our start pose robustness metric. We ran all simulations on a 3.7GHz 20-thread Intel Core i9-10900K CPU with 16GB of RAM. 
We used the following steerable needle hardware parameters in this simulation: the steerable needle's diameter was $d_\text{needle} = 1.0$ mm, the maximum insertion length was $l_\text{needle} = 150$ mm, and we varied the minimum radius of curvature within the interval $r_\text{min} = [50,100]$ mm depending on the scenario, which is within range of our physical steerable needle lung robot \cite{rox2020laser}. We also vary the insertion angle deviation $\theta$, which is the maximum deviation from the surface normal at the start position. Changing this parameter reflects the ability of the physician to reach a position on the insertion surface from a specific orientation, e.g., inside the airways, and to insert the steerable needle through the surface at this angle as discussed in Sec.~\ref{sec:surface}. The difficulty of such an insertion depends on the specific deployment scenario, e.g., inside the airways. 
\subsection{Sphere Scenario}
\label{sec:spheres}
\begin{figure}[t]
 \centering
 \includegraphics[width=1.0\columnwidth]{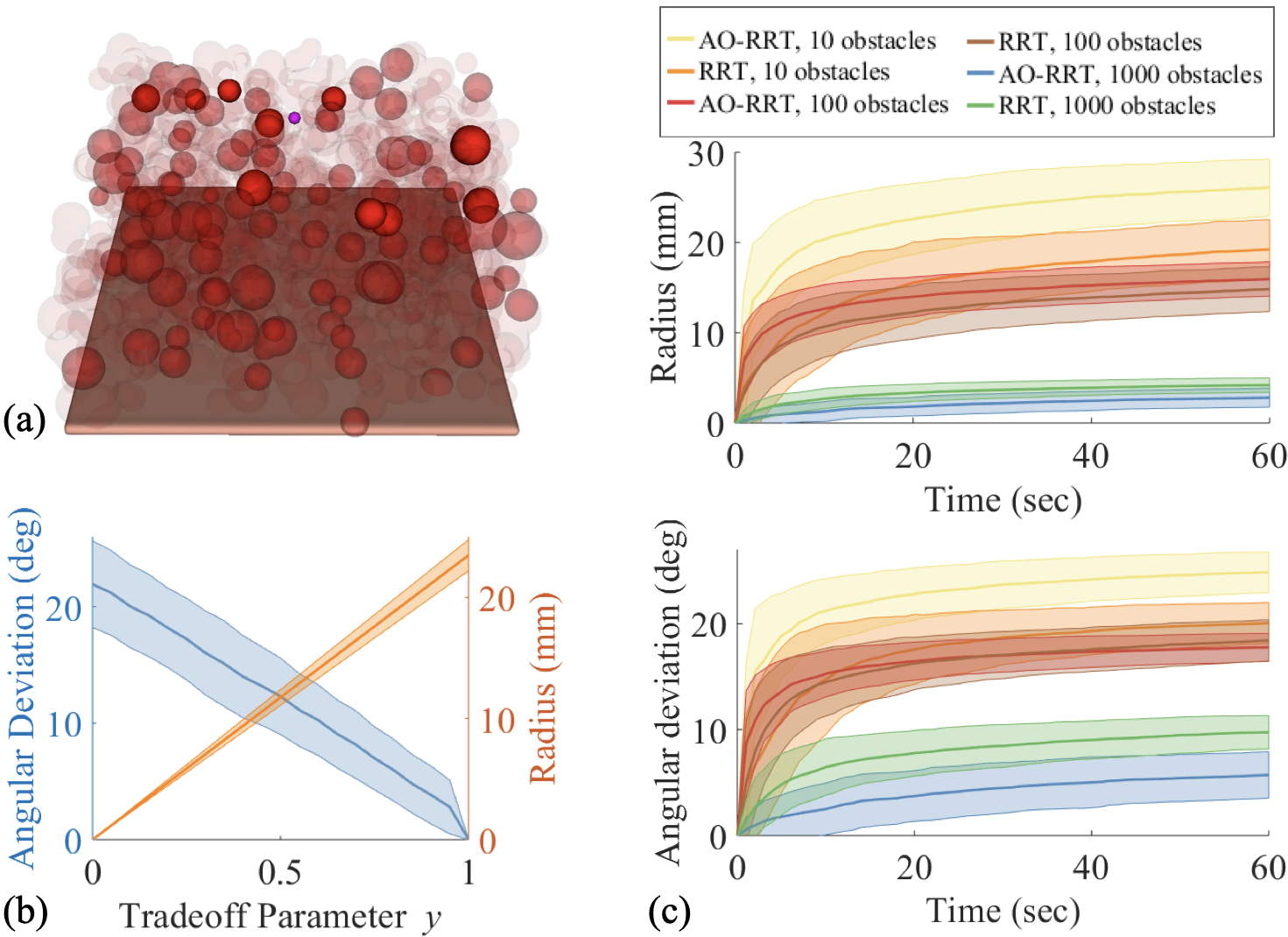}
 \caption{(a) Planning scenario with a planar start region (tan), a target (purple) and spherical obstacles (red). Levels of transparency for the spheres represent distinct scenarios with varying numbers of obstacles. (b) The tradeoff between robustness to changes in position and orientation for the plan with the largest safe start region radius found by the AO-RRT algorithm in a scenario with 10 obstacle spheres. (c) Largest safe start region radii and angular deviations found over time averaged over 50 random obstacle scenarios and with a standard deviation across 10 runs each for both the NeedleRRT and the AO-RRT algorithms. }
 \label{fig:aorrt_spheres}
 \end{figure}

We first tested our start pose robustness metric in an abstract scenario consisting of a planar start region and spherical obstacles with a fixed target position, as shown in Fig.~\ref{fig:aorrt_spheres} (a). This setup allowed us to systematically test our metric's behavior when faced with plans created by different planning algorithms and different levels of difficulty of the planning problem (by varying the number of obstacles and the size of the start region), as well as comparing our metric to a sampling-based heuristic.

We defined poses for the start region as positions located on the planar surface combined with orientations that deviated from the surface normal by up to $\theta = \pi/2$, i.e.~any, start orientation on the target-facing side of the surface was acceptable. 
We set the minimum radius of curvature of the steerable needle to $r_\text{min} = 100$ mm, 
and we defined the target position to be located $100$ mm above the center of the start region. 
We uniformly sampled random center positions and radii for the spherical obstacles, with a maximum radius of $10$ mm and a minimum distance of $20$ mm between the sphere centers and the fixed target.
For each combination of parameters tested in this setup, we created 50 planning environments consisting of randomly sampled obstacle spheres between the start region and the target position.

 The steerable needle motion planners we use throughout these simulations originally plan from a specific start pose to a target position. Therefore, we modified them to support backward planning from a target position towards the start region represented by a region of start poses $\Sigma$. This requires sampling multiple start orientations (as the target orientation is not fixed) as well as changing the target condition from accepting one position to multiple positions and orientations, as described in \cite{hoelscher2021backward}.

\subsubsection{Number of obstacles}
For our first simulation, we set the size of the start region to be $200 \times 200$ mm, and we varied the number of obstacle spheres from 10 obstacles to 1000 obstacles, as shown in Fig.~\ref{fig:aorrt_spheres} (a). To show that our start pose robustness metric is independent of the planner used, we applied two different steerable needle motion planners to this scenario. The first one is the NeedleRRT steerable needle planner by Patil et al.~\cite{patil2010interactive}, and the second one is an Asymptotic Optimality RRT (AO-RRT) \cite{hauser2016asymptotically} adapted to steerable needles.

We ran both planners for 60 seconds to find nominal motion plans and applied our metric to each plan found. For each of the 50 random sphere scenarios, we repeated this process 10 times. The planning algorithms had no knowledge of the metric. Fig.~\ref{fig:aorrt_spheres} (c) shows the best plans found over time with a maximum robustness to changes in position (tradeoff parameter $y=0$) and in orientation ($y=1$), respectively. The graphs visualize the average across the 50 random scenarios and 10 planning runs each, and the standard deviation across the 10 runs tracked over time. It can be seen that the number of obstacles plays a significant role in the size of the safe start region found, with more obstacles leading to significantly smaller safe start regions. This result is consistent across both the NeedleRRT and the AO-RRT. 

As outlined in Sec.~\ref{sec:surface}, there is a tradeoff between maximizing robustness to deviation in position and in orientation. Fig.~\ref{fig:aorrt_spheres} (b) visualizes this tradeoff for the plan with the largest safe start region radius found after 60 seconds in a scenario with 10 random obstacle spheres and using the AO-RRT planner, which is the best-performing combination of motion planner and number of obstacles in the setup described above. The tradeoff parameter $y$ is defined to be proportional to the radius $\rho$ of the safe start region our metric determines, and the planner found plans with a maximum radius of 23.5 mm on average  (standard deviation 1.3 mm). It can be seen that as the value of $y$ increases, the angular deviation $\alpha$ decreases almost linearly from an initial value of $21.9^\circ$ degrees to zero degrees.

\subsubsection{Start region size}
\begin{figure}[t]
 \centering
 \includegraphics[width=1.0\columnwidth]{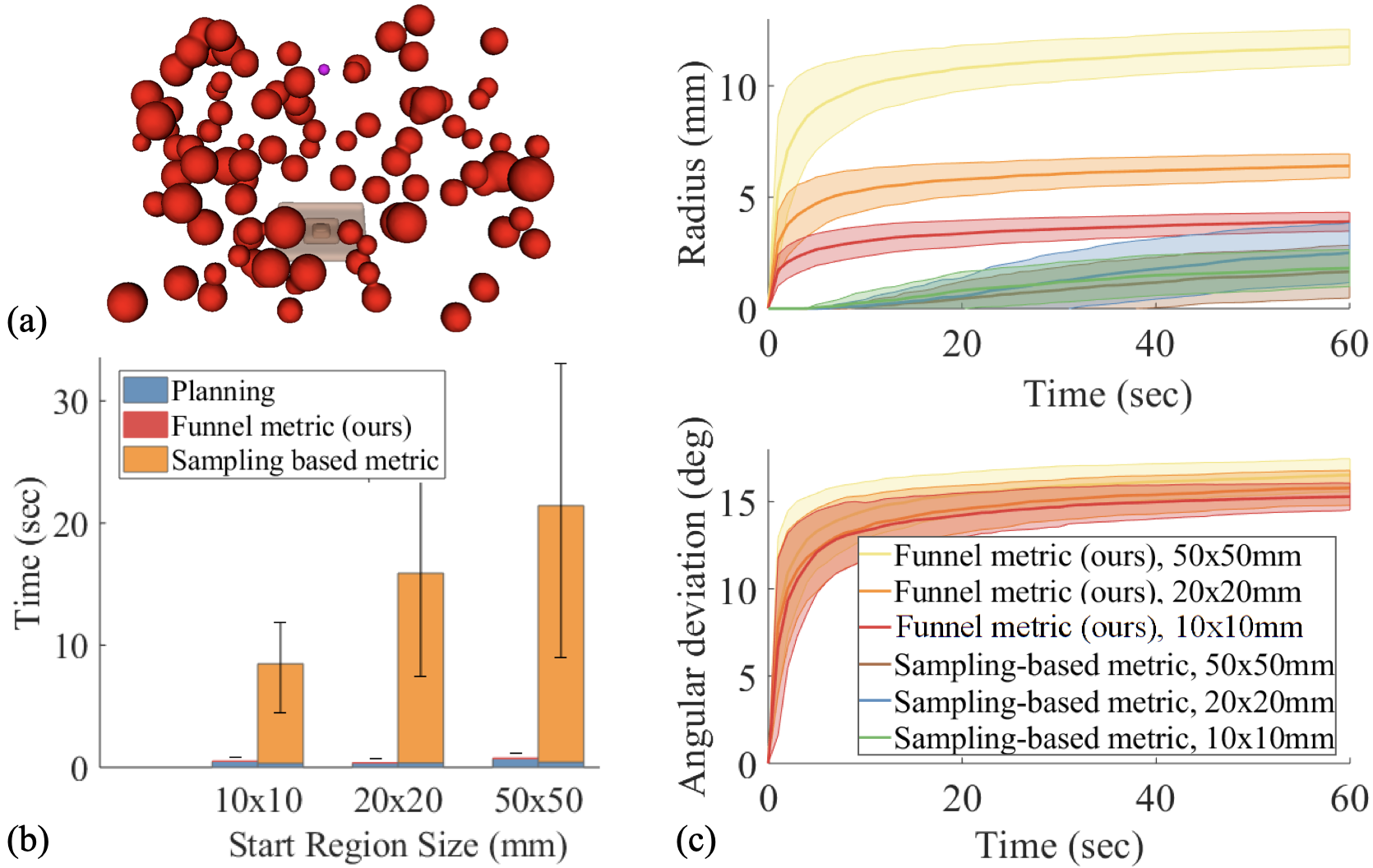}
 \caption{(a) Planning scenario with a planar start region (tan), a target (purple) and spherical obstacles (red). Levels of transparency represent scenarios with a varying size of the start region. (b) Our metric is robust to changes in the start region size; computing safe start regions significantly faster than a sampling-based heuristic. (c) Our metric computes safe start regions faster and more accurately, resulting in larger radii and angular deviations found over time compared to a sampling-based metric, which only computes radii.}
 \label{fig:startSize}
 \end{figure}

In our second set of experiments, we tested our start pose robustness metric using different sizes for the start region, varying from $10 \times 10$ mm to $50 \times 50$ mm, and a fixed number of 100 obstacle spheres, as shown in Fig.~\ref{fig:startSize} (a). We also compared our metric with a baseline sampling-based robustness heuristic. We used the NeedleRRT planning algorithm for all planning runs in this simulation.
 
 For the sampling-based heuristic, we ran the steerable needle motion planner the same way as for our start pose robustness metric to create a nominal motion plan, with the motion planner being ignorant of the plan evaluation strategy used.  
 When a nominal plan was found after $t_i$ seconds, we re-ran the steerable needle motion planner for the sampling-based heuristic computation. This time it did not terminate after one path from the target to the start region was found but extended the search tree with the goal of reaching all triangles representing the start region. To speed up this process for a fair comparison, we included goal biasing with a probability of $10\%$ per sample towards triangles closer to the nominal start pose than the pose's minimum distance to an obstacle. 
 We also pruned samples in the search tree from which the start region could not be reached.
 We terminated the tree extension process when either plans to all eligible triangles were found or if no new plan to a previously not reached triangle was found in the past $2 t_i$ seconds. Then, we determined the distance between the nominal start pose and the closest triangle to which no plan from the target was found, which limits the size of the determined safe start region. Note that while our new funnel method can be used with any steerable needle motion planner, this comparison method relies on iteratively building a tree structure that can eventually reach multiple targets, such as in RRT algorithms.

 Our second experiment consisted of running the planning process 10 times for 60 seconds each and repeating this procedure for 50 scenarios consisting of randomly sampled obstacles. Fig.~\ref{fig:startSize} (b) shows how much of the 60 seconds is spent by the motion planner to find nominal motion plans and by the two evaluation strategies to compute the metrics. The motion planner takes, on average, 0.5 sec to find a motion plan, which is independent of the plan evaluation. Our new metric computes a safe start region in, on average, 0.03 sec. This value is consistent across different start region sizes. In contrast, the duration of the sampling-based heuristic is significantly higher than that of our new metric, and it increases with the start region size. For a start region size of $50 \times 50$ mm, the sampling-based heuristic takes 21 seconds, which is 700 times slower than our metric.

Fig.~\ref{fig:startSize} (c) shows the average best plan found in terms of position deviation and orientation deviation tracked over time. 
 For position deviation, our metric not only determines plans with larger safe start regions faster but also finds larger safe start regions overall. Our start pose robustness metric finds safe start regions whose size in position grows with the size of the overall start region, from 3.9 mm (0.4 mm standard deviation) for the $10 \times 10$ mm start region to 11.7 mm (0.7 mm standard deviation) for the $50 \times 50$ mm start region. For the sampling-based metric, the largest radius found for the $50 \times 50$ mm start region is 2.5 mm (1.3 mm standard deviation). As the sampling-based heuristic only computes single start orientations and no orientation ranges, its angular deviation is zero and is not shown. In angular deviation, the change between these sizes is much less pronounced, from $15.3^\circ$ ($0.8^\circ$s standard deviation) to $16.8^\circ$ ($1.4^\circ$ standard deviation). Unlike the position deviation, the allowable angular deviation is not directly influenced by the size of the start region, only indirectly by the reduced choice of plans in the case of a smaller start region.
\subsection{Liver Scenario}
\label{sec:liver}

\begin{figure*}[t]
 \centering
 \includegraphics[width=1.0\textwidth]{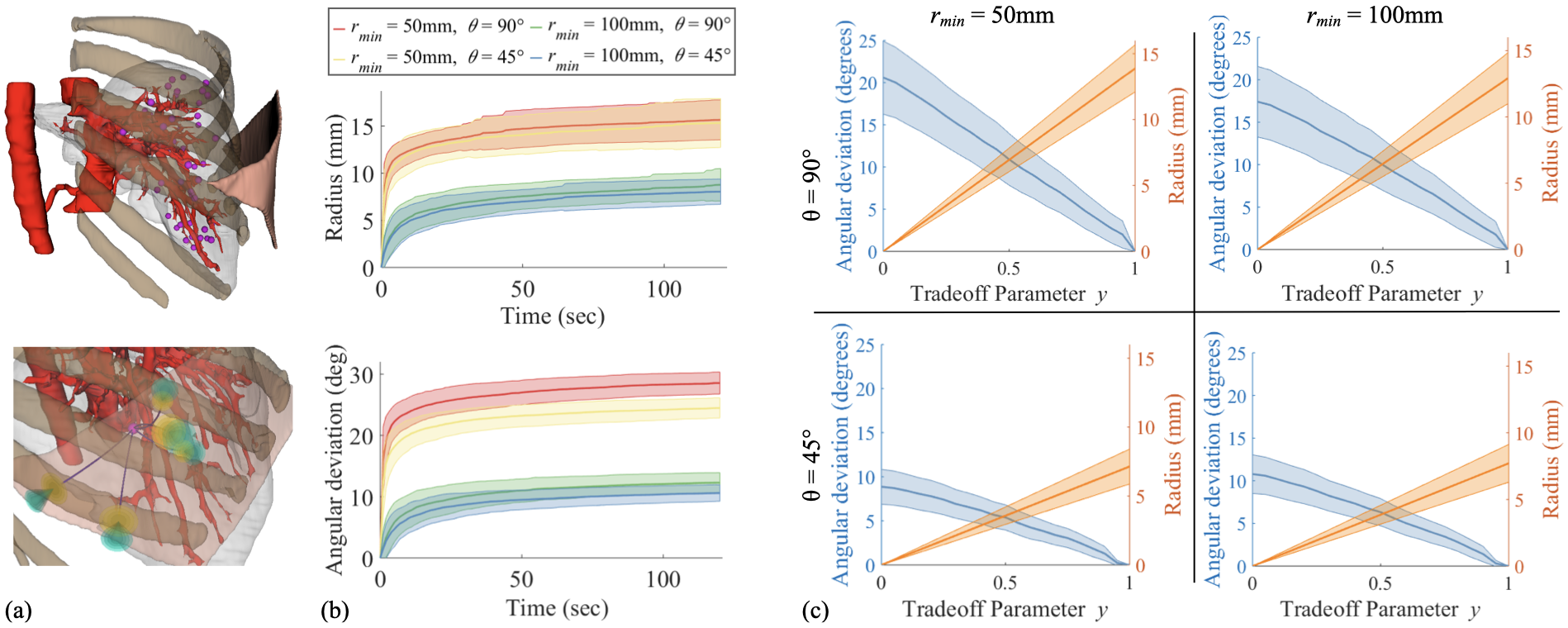}
 \caption{(a) Liver procedure scenario with a percutaneous start region (tan) and a steerable needle navigating towards 50 randomly sampled targets (pink). Critical obstacles to be avoided include blood vessels (red), ribs (brown), and the outside of the liver tissue (grey) once the needle tip has entered the liver. Our method identifies safe deviations from the start pose of nominal motion plans (blue) in both position (yellow) and orientation (cyan). (b) Largest safe start region radii and angular deviations found over time averaged over 50 random obstacle scenarios and with a standard deviation across 10 runs, each with different values for the steerable needle's minimum radius of curvature $r_\text{min}$ and surface angle $s$.
 (c) Trade-off between robustness to changes in position and orientation for the plan with the largest safe start region radius found in 2 min with different parameters for $r_\text{min}$ and $s$.}
 \label{fig:liver}
 \end{figure*}

Liver cancer accounts for 
over 30,000 deaths per year in the United States \cite{cancerSociety} and it is one of the few types of cancer whose incidence worldwide is still growing \cite{bray2018global}. Early diagnosis is important for increasing patient survival, and needle biopsy of suspicious nodules is commonly used for definitive diagnosis~\cite{torbenson2015liver}. During these procedures, the needle is inserted percutaneously between the ribs and into the liver~\cite{bodard2023percutaneous}. Several works have proposed using steerable needles during this procedure to enable safe access to otherwise hard-to-reach targets~\cite{adebar2015methods, Reed2011_RAM, liu2016fast}.  

Our liver planning scenario is based on a CT scan from The Cancer Imaging Archive (TCIA)\cite{clark2013cancer} and segmentation of critical anatomy from this CT scan published in \cite{fried2022clinical}. We identified an area of the skin in the right upper abdomen as the start region and converted it into a 3D mesh using 3D Slicer \cite{kikinis20143d}.
We defined liver tissue and the tissue between the start region and the liver as the steerable needle's workspace, and we classified large blood vessels and ribs as critical obstacles that had to be avoided. A visualization of the planning scenario is shown in Fig.\ref{fig:liver} (a). We sampled 50 random targets inside the liver with a minimum distance to critical obstacles of 10 mm and a minimum distance to the start region of 50 mm. We used our backward steerable needle planner \cite{hoelscher2021backward} to find motion plans from the target to the the start. Examples of resulting plans can be seen in Fig.~\ref{fig:liver} (a).
In this simulation, we varied two parameters. We chose the steerable needle's minimum radius of curvature to be $r_\text{min} = [50,100]$ mm, which are representative values depending on the needle design \cite{rox2020laser}. 
We set the maximum insertion angle to $\theta = [45,90]^\circ$, which is the maximum deviation from the start region normal direction.
We ran the planner for 120 seconds for each of the targets, and we repeated this process 10 times per target. The results of this simulation shown in Fig.\ref{fig:liver} (b) show the best plans found over time in terms of position robustness and orientation robustness averaged across all 50 targets and 10 runs and the average standard deviation across the runs.  
It can be seen that changing the maximum insertion angle $\theta$ barely makes a difference in position robustness and only a small difference in orientation robustness. However, a change of $r_\text{min}$ from 100 mm to 50 mm almost doubles the position size of the best safe start region from 8.8 mm to 15.7 mm radius, and more than doubles the orientation deviation from $12.3^\circ$ to $28.5^\circ$. Analyzing the influence of tradeoff parameter $y$ as depicted in Fig.~\ref{fig:liver} (c) leads to similar results, showing that the steerable needle's minimum radius of curvature has a much higher influence than the maximum insertion angle on the size of the safe start region in both position and orientation.

\subsection{Lung Scenario}
\label{sec:lung}
Lung cancer is responsible for the most cancer-related deaths in the United States \cite{cancerSociety} and early diagnosis through biopsy can increase a patient's chances of survival.
 A novel approach for safe lung nodule biopsy involves a physician navigating a bronchoscope transorally through the airways and piercing a steerable needle through the airway wall into the lung parenchyma. At this point, the physician hands off control to a robot that autonomously steers the needle to the target \cite{kuntz2016toward, kuntz2023autonomous}.
Using a steerable needle in a biopsy procedure can potentially mitigate patient risk, as the needle is able to avoid collisions with critical obstacles.

\begin{figure*}[t]
 \centering
 \includegraphics[width=1.0\textwidth]{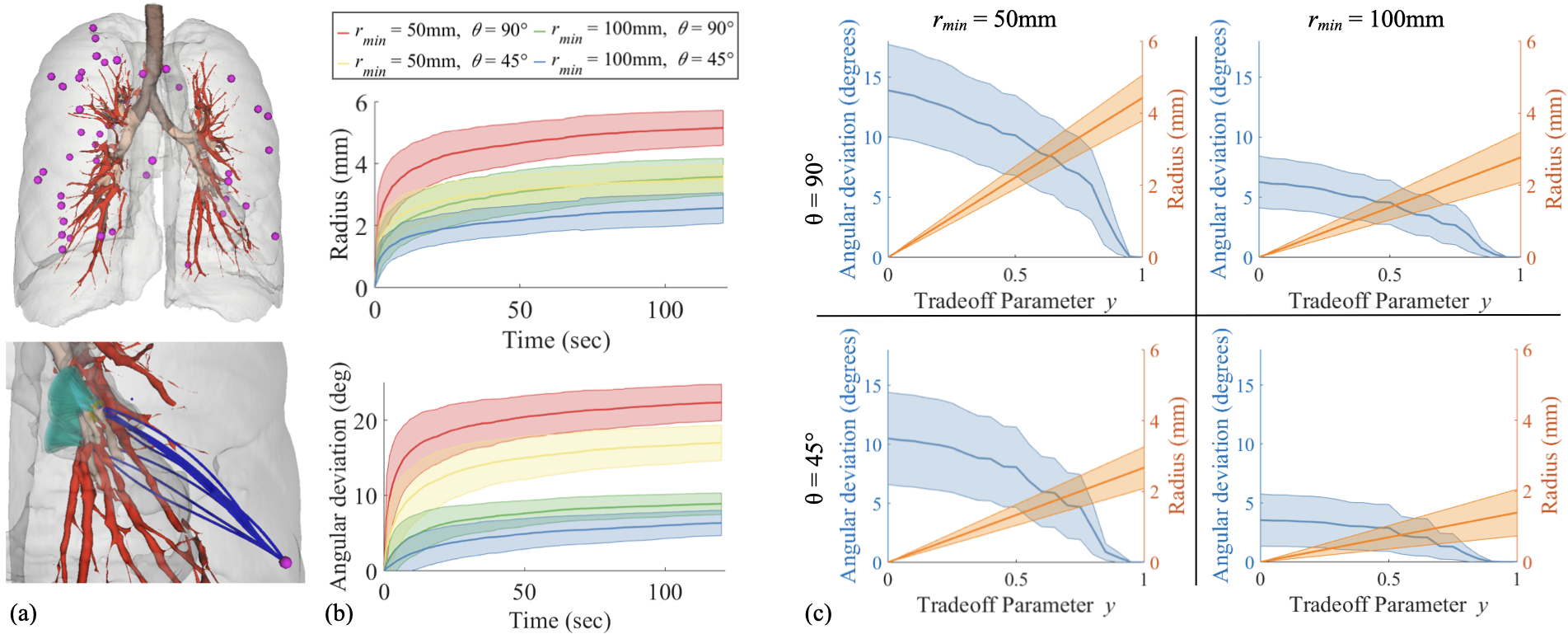}
 \caption{(a) Lung procedure scenario with a transoral approach planning from the airways (tan) towards 50 random targets (pink). Light tan airways represent suitable start regions reachable by a bronchoscope. Critical obstacles to be avoided include blood vessels (red) and the outside of the lung tissue (grey). Our method identifies safe deviations from the start poses of nominal motion plans (blue) in both position (yellow) and orientation (cyan). (b) Largest safe start region radii and angular deviations found over time averaged over 50 random obstacle scenarios and with a standard deviation across 10 runs each with different values for the steerable needle's minimum radius of curvature $r_\text{min}$ and surface angle $s$.
 (c) Trade-off between robustness to changes in position and orientation for the plan with the largest safe start region radius found in 2 min with different parameters for $r_\text{min}$ and $s$.}
 \label{fig:lung}
 \end{figure*}

To model this scenario, we used a CT scan of in-vivo human lungs from the Lung Image Database Consortium and Image Database Resource Initiative (LIDC-IDRI) image collection \cite{armato2011lung} with a voxel size of $(0.6 \times 0.6 \times 0.7)$mm.
We segmented the anatomy relevant for the planning environment using the method described in \cite{fu2018safe, fried2023dataset}.
We used the surface of the segmented airways as the start region for the steerable needle and represented it as a 3D mesh. We classified major blood vessels and the boundary of the lung parenchyma as obstacles. We then randomly sampled $50$ target positions in the lung with a minimum distance of $40$mm to the airways and a minimum distance of 10 mm to critical obstacles. The lung environment as well as the targets are shown in Fig.~\ref{fig:lung} (a).
 
 The setup of this simulation is very similar to the liver scenario running our RRT needle planner \cite{hoelscher2021backward} for up to 120 seconds for 50 random targets, and repeating this process for 10 runs per target. Again, we analyze the influence of changing the steerable needle's minimum radius of curvature $r_\text{min}$ and the maximum insertion angle $\theta$ using the same range of values as in the liver scenario. Results are shown in Fig.~\ref{fig:lung} (b)-(c).
 One of the main differences between the liver scenario and the airway scenario is the nature of the start region. The airways are relatively narrow, which results in comparatively low position robustness found over time.  The best combination of parameters ($r_\text{min}=100, \theta = 90^\circ$) resulted in a 5.2 mm radius (0.5 mm standard deviation) found after 120 seconds, which is less than a third of the radius found in the liver scenario (see Fig.~\ref{fig:lung} (b)).
 The robustness to changes in orientation, however, is $22.3^\circ$ ($2.5^\circ$ standard deviation), which is less than 25\% smaller than that found in the liver scenario.
 
 Furthermore, due to the airways' cylindrical shape, the start region surface is curving away from the original start pose, which reduces the amount of overlap between the orientation ranges found by our method and the physician start ranges. This results in a more pronounced change in the size of the safe start region when we vary the maximum insertion angle $\theta$. This can be seen in Fig.~\ref{fig:lung} (c), where for all scenarios, the angular deviation drops increasingly fast with a growing trade-off parameter $y$. Overall, the safe start regions in the lung scenario are smaller in both position and orientation, making this a more challenging scenario for the physician deploying the steerable needle.
\section{Conclusion}
In this work, we introduced a start pose robustness metric for steerable needle motion plans that determines the robustness to deviations from a nominal start pose in both position and orientation. 
Our metric can be applied to any steerable needle planning scenario with motion plans starting on a needle insertion surface. The metric computes the size of a safe start region consisting of safe start poses from which a collision-free motion plan to the target exists. Our method takes into account the tradeoff between maximizing the robustness to deviations in position and in orientation. Our method allows the physician to adapt this tradeoff based on their preference and the requirements of the specific scenario by the simple change of a parameter.
Our simulations demonstrated that our metric can be combined with different steerable needle motion planners and that it was significantly faster than a sampling-based comparison heuristic. We showed its performance in two medical scenarios, a liver biopsy scenario and a transoral lung access scenario, under different hardware constraints. Furthermore, we evaluated how different values for the tradeoff parameter influence the resulting safe start regions.

In future work, we will consider using the orientation range data structure developed in this work not only to analyze the start pose of the steerable needle but also to track its deployment along and possible deviation from a nominal motion plan. The computed orientation ranges could be used to determine how much the needle can deviate from the plan, both in position and orientation and still recover and reach the target.
Moreover, in the current setup, the motion planners used to create nominal motion plans are agnostic to the start pose robustness metric. Instead of hoping to find better plans over time, we plan to integrate the metric into a steerable needle motion planner that systematically searches for better plans with respect to this metric.
Most importantly, we aim to test our new start pose robustness metric and its visualization in a physical experiment in which a physician can customize the tradeoff parameter to match their preferences.

\section*{ACKNOWLEDGMENT}

The authors would like to acknowledge the National Cancer Institute and the Foundation for the National Institutes of Health, and their critical role in the creation of the free publicly available LIDC/IDRI Database used in this study. The authors also would like to thank Alan Kuntz, Mengyu Fu, Margaret Rox, Maxwell Emerson, Tayfun Ertop, Fabien Maldonado, Erin Gillaspie, and Yueh Lee for their valuable insights and feedback.

\bibliographystyle{IEEEtran}
\bibliography{main.bib}

\begin{thebibliography}{10}
\providecommand{\url}[1]{#1}
\csname url@samestyle\endcsname
\providecommand{\newblock}{\relax}
\providecommand{\bibinfo}[2]{#2}
\providecommand{\BIBentrySTDinterwordspacing}{\spaceskip=0pt\relax}
\providecommand{\BIBentryALTinterwordstretchfactor}{4}
\providecommand{\BIBentryALTinterwordspacing}{\spaceskip=\fontdimen2\font plus
\BIBentryALTinterwordstretchfactor\fontdimen3\font minus \fontdimen4\font\relax}
\providecommand{\BIBforeignlanguage}[2]{{%
\expandafter\ifx\csname l@#1\endcsname\relax
\typeout{** WARNING: IEEEtran.bst: No hyphenation pattern has been}%
\typeout{** loaded for the language `#1'. Using the pattern for}%
\typeout{** the default language instead.}%
\else
\language=\csname l@#1\endcsname
\fi
#2}}
\providecommand{\BIBdecl}{\relax}
\BIBdecl

\bibitem{adebar2015methods}
T.~K. Adebar, J.~D. Greer, P.~F. Laeseke, G.~L. Hwang, and A.~M. Okamura, ``Methods for improving the curvature of steerable needles in biological tissue,'' \emph{IEEE Transactions on Biomedical Engineering}, vol.~63, no.~6, pp. 1167--1177, 2015.

\bibitem{Secoli2022_PLOSOne}
\BIBentryALTinterwordspacing
R.~Secoli, E.~Matheson, M.~Pinzi, S.~Galvan, A.~Donder, T.~Watts, M.~Riva, D.~D. Zani, L.~Bello, and F.~Rodriguez~y Baena, ``Modular robotic platform for precision neurosurgery with a bio-inspired needle: System overview and first in-vivo deployment,'' \emph{PLOS ONE}, vol.~17, no.~10, pp. 1--29, 10 2022. [Online]. Available: \url{https://doi.org/10.1371/journal.pone.0275686}
\BIBentrySTDinterwordspacing

\bibitem{kuntz2023autonomous}
A.~Kuntz, M.~Emerson, T.~E. Ertop, I.~Fried, M.~Fu, J.~Hoelscher, M.~Rox, J.~Akulian, E.~A. Gillaspie, Y.~Z. Lee \emph{et~al.}, ``Autonomous medical needle steering in vivo,'' \emph{Science Robotics}, vol.~8, no.~82, p. eadf7614, 2023.

\bibitem{majewicz2013cartesian}
A.~Majewicz and A.~M. Okamura, ``Cartesian and joint space teleoperation for nonholonomic steerable needles,'' in \emph{2013 World Haptics Conference (WHC)}.\hskip 1em plus 0.5em minus 0.4em\relax IEEE, 2013, pp. 395--400.

\bibitem{hoelscher2021backward}
J.~Hoelscher, M.~Fu, I.~Fried, M.~Emerson, T.~E. Ertop, M.~Rox, A.~Kuntz, J.~A. Akulian, R.~J. Webster~III, and R.~Alterovitz, ``Backward planning for a multi-stage steerable needle lung robot,'' \emph{Robotics and Automation Letters (RA-L)}, vol.~6, no.~2, pp. 3987--3994, 2021.

\bibitem{patil2014needle}
S.~Patil, J.~Burgner, R.~J. Webster, and R.~Alterovitz, ``Needle steering in 3-d via rapid replanning,'' \emph{IEEE Transactions on Robotics}, vol.~30, no.~4, pp. 853--864, 2014.

\bibitem{webster2006nonholonomic}
R.~J. Webster~III, J.~S. Kim, N.~J. Cowan, G.~S. Chirikjian, and A.~M. Okamura, ``Nonholonomic modeling of needle steering,'' \emph{The International Journal of Robotics Research}, vol.~25, no. 5-6, pp. 509--525, 2006.

\bibitem{hauser2016asymptotically}
K.~Hauser and Y.~Zhou, ``Asymptotically optimal planning by feasible kinodynamic planning in a state--cost space,'' \emph{IEEE Transactions on Robotics}, vol.~32, no.~6, pp. 1431--1443, 2016.

\bibitem{cowan2011robotic}
N.~J. Cowan, K.~Goldberg, G.~S. Chirikjian, G.~Fichtinger, R.~Alterovitz, K.~B. Reed, V.~Kallem, W.~Park, S.~Misra, and A.~M. Okamura, ``Robotic needle steering: Design, modeling, planning, and image guidance,'' in \emph{Surgical Robotics}.\hskip 1em plus 0.5em minus 0.4em\relax Springer, 2011, pp. 557--582.

\bibitem{rox2020laser}
M.~{Rox}, M.~{Emerson}, T.~E. {Ertop}, I.~{Fried}, M.~{Fu}, J.~{Hoelscher}, A.~{Kuntz}, J.~{Granna}, J.~{Mitchell}, M.~{Lester}, F.~{Maldonado}, E.~A. {Gillaspie}, J.~A. {Akulian}, R.~{Alterovitz}, and R.~J. {Webster}, ``Decoupling steerability from diameter: Helical dovetail laser patterning for steerable needles,'' \emph{IEEE Access}, vol.~8, pp. 181\,411--181\,419, 2020.

\bibitem{minhas2007modeling}
D.~S. Minhas, J.~A. Engh, M.~M. Fenske, and C.~N. Riviere, ``Modeling of needle steering via duty-cycled spinning,'' in \emph{2007 29th Annual International Conference of the IEEE Engineering in Medicine and Biology Society}.\hskip 1em plus 0.5em minus 0.4em\relax IEEE, 2007, pp. 2756--2759.

\bibitem{rucker2013sliding}
D.~C. Rucker, J.~Das, H.~B. Gilbert, P.~J. Swaney, M.~I. Miga, N.~Sarkar, and R.~J. Webster, ``Sliding mode control of steerable needles,'' \emph{IEEE Transactions on Robotics}, vol.~29, no.~5, pp. 1289--1299, 2013.

\bibitem{ertop2020steerable}
T.~E. Ertop, M.~Emerson, M.~Rox, J.~Granna, R.~Webster, F.~Maldonado, E.~Gillaspie, M.~Lester, A.~Kuntz, C.~Rucker \emph{et~al.}, ``Steerable needle trajectory following in the lung: Torsional deadband compensation and full pose estimation with 5dof feedback for needles passing through flexible endoscopes,'' in \emph{Dynamic Systems and Control Conference}, vol. 84270.\hskip 1em plus 0.5em minus 0.4em\relax American Society of Mechanical Engineers, 2020, p. V001T05A003.

\bibitem{fried2021design}
I.~Fried, J.~Hoelscher, M.~Fu, M.~Emerson, T.~E. Ertop, M.~Rox, J.~Granna, A.~Kuntz, J.~A. Akulian, R.~J. Webster \emph{et~al.}, ``Design considerations for a steerable needle robot to maximize reachable lung volume,'' in \emph{2021 IEEE International Conference on Robotics and Automation (ICRA)}.\hskip 1em plus 0.5em minus 0.4em\relax IEEE, 2021, pp. 1418--1425.

\bibitem{duindam2010three}
V.~Duindam, J.~Xu, R.~Alterovitz, S.~Sastry, and K.~Goldberg, ``Three-dimensional motion planning algorithms for steerable needles using inverse kinematics,'' \emph{The International Journal of Robotics Research}, vol.~29, no.~7, pp. 789--800, 2010.

\bibitem{bano2012smooth}
S.~Bano, S.~Y. Ko, and F.~R. y~Baena, ``Smooth path planning for a biologically-inspired neurosurgical probe,'' in \emph{2012 Annual International Conference of the IEEE Engineering in Medicine and Biology Society}.\hskip 1em plus 0.5em minus 0.4em\relax IEEE, 2012, pp. 920--923.

\bibitem{segato2021optimized}
A.~Segato, V.~Corbetta, J.~Zangari, S.~Perri, F.~Calimeri, and E.~De~Momi, ``Optimized 3d path planner for steerable catheters with deductive reasoning,'' in \emph{2021 IEEE International Conference on Robotics and Automation (ICRA)}.\hskip 1em plus 0.5em minus 0.4em\relax IEEE, 2021, pp. 1466--1472.

\bibitem{li2014combination}
P.~Li, S.~Jiang, J.~Yang, and Z.~Yang, ``A combination method of artificial potential field and improved conjugate gradient for trajectory planning for needle insertion into soft tissue,'' \emph{Journal of Medical and Biological Engineering}, vol.~34, no.~6, pp. 568--573, 2014.

\bibitem{caborni2012risk}
C.~Caborni, S.~Y. Ko, E.~De~Momi, G.~Ferrigno, and F.~R. y~Baena, ``Risk-based path planning for a steerable flexible probe for neurosurgical intervention,'' in \emph{2012 4th IEEE RAS \& EMBS International Conference on Biomedical Robotics and Biomechatronics (BioRob)}.\hskip 1em plus 0.5em minus 0.4em\relax IEEE, 2012, pp. 866--871.

\bibitem{xu2008motion}
J.~Xu, V.~Duindam, R.~Alterovitz, and K.~Goldberg, ``Motion planning for steerable needles in 3d environments with obstacles using rapidly-exploring random trees and backchaining,'' in \emph{2008 IEEE international conference on automation science and engineering}.\hskip 1em plus 0.5em minus 0.4em\relax IEEE, 2008, pp. 41--46.

\bibitem{sun2015stochastic}
W.~Sun, J.~Van Den~Berg, and R.~Alterovitz, ``Stochastic extended {LQR}: Optimization-based motion planning under uncertainty,'' in \emph{Algorithmic Foundations of Robotics XI}.\hskip 1em plus 0.5em minus 0.4em\relax Springer, 2015, pp. 609--626.

\bibitem{lavalle1998rapidly}
S.~M. LaValle, ``Rapidly-exploring random trees: A new tool for path planning,'' \emph{TR 98-11, Computer Science Dept., Iowa State Univ.}, 1998.

\bibitem{patil2010interactive}
S.~Patil and R.~Alterovitz, ``Interactive motion planning for steerable needles in 3d environments with obstacles,'' in \emph{Proc. IEEE RAS \& EMBS International Conference on Biomedical Robotics and Biomechatronics}, 2010, pp. 893--899.

\bibitem{favaro2021evolutionary}
A.~Favaro, A.~Segato, F.~Muretti, and E.~De~Momi, ``An evolutionary-optimized surgical path planner for a programmable bevel-tip needle,'' \emph{IEEE Transactions on Robotics}, vol.~37, no.~4, pp. 1039--1050, 2021.

\bibitem{liu2016fast}
F.~Liu, A.~Garriga-Casanovas, R.~Secoli, and F.~\{Rodriguez y Baena\}, ``Fast and adaptive fractal tree-based path planning for programmable bevel tip steerable needles,'' \emph{IEEE Robotics and Automation Letters}, vol.~1, no.~2, pp. 601--608, 2016.

\bibitem{pinzi2019adaptive}
M.~Pinzi, S.~Galvan, and F.~{Rodriguez y Baena}, ``The adaptive hermite fractal tree ({AHFT}): a novel surgical 3d path planning approach with curvature and heading constraints,'' \emph{International Journal of Computer Assisted Radiology and Surgery}, vol.~14, no.~4, pp. 659--670, 2019.

\bibitem{fu2023certifiable}
M.~Fu, K.~Solovey, O.~Salzman, and R.~Alterovitz, ``Toward certifiable optimal motion planning for medical steerable needles,'' \emph{The International Journal of Robotics Research}, vol.~42, no.~10, pp. 798--826, 2023.

\bibitem{segato2021inverse}
A.~Segato, M.~Di~Marzo, S.~Zucchelli, S.~Galvan, R.~Secoli, and E.~De~Momi, ``Inverse reinforcement learning intra-operative path planning for steerable needle,'' \emph{IEEE Transactions on Biomedical Engineering}, vol.~69, no.~6, pp. 1995--2005, 2021.

\bibitem{sun2015replanning}
W.~Sun, S.~Patil, and R.~Alterovitz, ``High-frequency replanning under uncertainty using parallel sampling-based motion planning,'' \emph{IEEE Transactions on Robotics}, vol.~31, no.~1, pp. 104--116, 2015.

\bibitem{pinzi2021replanning}
M.~Pinzi, T.~Watts, R.~Secoli, S.~Galvan, and F.~{Rodriguez y Baena}, ``Path replanning for orientation-constrained needle steering,'' \emph{IEEE Transactions on Biomedical Engineering}, vol.~68, no.~5, pp. 1459--1466, 2021.

\bibitem{segato2022hybrid}
A.~Segato, F.~Calimeri, I.~Testa, V.~Corbetta, M.~Riva, and E.~De~Momi, ``A hybrid inductive learning-based and deductive reasoning-based 3-d path planning method in complex environments,'' \emph{Autonomous Robots}, vol.~46, no.~5, pp. 645--666, 2022.

\bibitem{kuntz2015motion}
A.~Kuntz, L.~G. Torres, R.~H. Feins, R.~J. Webster, and R.~Alterovitz, ``Motion planning for a three-stage multilumen transoral lung access system,'' in \emph{Proc. IEEE/RSJ International Conference on Intelligent Robots and Systems (IROS)}, 2015, pp. 3255--3261.

\bibitem{alterovitz2008motion}
R.~Alterovitz, M.~Branicky, and K.~Goldberg, ``Motion planning under uncertainty for image-guided medical needle steering,'' \emph{The International Journal of Robotics Research}, vol.~27, no. 11-12, pp. 1361--1374, 2008.

\bibitem{park2005diffusion}
W.~Park, J.~S. Kim, Y.~Zhou, N.~J. Cowan, A.~M. Okamura, and G.~S. Chirikjian, ``Diffusion-based motion planning for a nonholonomic flexible needle model,'' in \emph{Proc. International Conference on Robotics and Automation (ICRA)}.\hskip 1em plus 0.5em minus 0.4em\relax IEEE, 2005, pp. 4600--4605.

\bibitem{van2012motion}
J.~{van den Berg}, S.~Patil, and R.~Alterovitz, ``Motion planning under uncertainty using iterative local optimization in belief space,'' \emph{The International Journal of Robotics Research}, vol.~31, no.~11, pp. 1263--1278, 2012.

\bibitem{alterovitz2005planning}
R.~Alterovitz, K.~Goldberg, and A.~Okamura, ``Planning for steerable bevel-tip needle insertion through 2d soft tissue with obstacles,'' in \emph{Proc. International Conference on Robotics and Automation (ICRA)}.\hskip 1em plus 0.5em minus 0.4em\relax IEEE, 2005, pp. 1640--1645.

\bibitem{patil2011motion}
S.~Patil, J.~van~den Berg, and R.~Alterovitz, ``Motion planning under uncertainty in highly deformable environments,'' in \emph{Proc. Robotics Science and Systems (RSS)}.\hskip 1em plus 0.5em minus 0.4em\relax RSS, 2011.

\bibitem{segato2021position}
A.~Segato, C.~Di~Vece, S.~Zucchelli, M.~Di~Marzo, T.~Wendler, M.~F. Azampour, S.~Galvan, R.~Secoli, and E.~De~Momi, ``Position-based dynamics simulator of brain deformations for path planning and intra-operative control in keyhole neurosurgery,'' \emph{IEEE Robotics and Automation Letters}, vol.~6, no.~3, pp. 6061--6067, 2021.

\bibitem{duindam2008screw}
V.~Duindam, R.~Alterovitz, S.~Sastry, and K.~Goldberg, ``Screw-based motion planning for bevel-tip flexible needles in 3d environments with obstacles,'' in \emph{2008 IEEE international conference on robotics and automation}.\hskip 1em plus 0.5em minus 0.4em\relax IEEE, 2008, pp. 2483--2488.

\bibitem{burrows2015smooth}
C.~Burrows, F.~Liu, and F.~R. y~Baena, ``Smooth on-line path planning for needle steering with non-linear constraints,'' in \emph{2015 IEEE/RSJ International Conference on Intelligent Robots and Systems (IROS)}.\hskip 1em plus 0.5em minus 0.4em\relax IEEE, 2015, pp. 2653--2658.

\bibitem{fu2018safe}
M.~Fu, A.~Kuntz, R.~J. Webster, and R.~Alterovitz, ``Safe motion planning for steerable needles using cost maps automatically extracted from pulmonary images,'' in \emph{Proc. IEEE/RSJ International Conference on Intelligent Robots and Systems (IROS)}.\hskip 1em plus 0.5em minus 0.4em\relax IEEE, 2018, pp. 4942--4949.

\bibitem{bentley2023safer}
M.~Bentley, C.~Rucker, C.~Reddy, O.~Salzman, and A.~Kuntz, ``Safer motion planning of steerable needles via a shaft-to-tissue force model,'' \emph{Journal of Medical Robotics Research}, 2023.

\bibitem{favaro2018uncertainty}
A.~Favaro, L.~Cerri, S.~Galvan, F.~{Rodriguez y Baena}, and E.~De~Momi, ``Automatic optimized 3d path planner for steerable catheters with heuristic search and uncertainty tolerance,'' in \emph{Proc. International Conference on Robotics and Automation (ICRA)}.\hskip 1em plus 0.5em minus 0.4em\relax IEEE, 2018, pp. 9--16.

\bibitem{hoelscher2022metric}
J.~Hoelscher, I.~Fried, M.~Fu, M.~Patwardhan, M.~Christman, J.~Akulian, R.~J. Webster, and R.~Alterovitz, ``A metric for finding robust start positions for medical steerable needle automation,'' in \emph{2022 IEEE/RSJ International Conference on Intelligent Robots and Systems (IROS)}.\hskip 1em plus 0.5em minus 0.4em\relax IEEE, 2022, pp. 9526--9533.

\bibitem{hong20193d}
A.~Hong, Q.~Boehler, R.~Moser, A.~Zemmar, L.~Stieglitz, and B.~J. Nelson, ``3d path planning for flexible needle steering in neurosurgery,'' \emph{The International Journal of Medical Robotics and Computer Assisted Surgery}, vol.~15, no.~4, p. e1998, 2019.

\bibitem{pinzi2021computer}
V.~Vakharia and F.~R. y~Baena, ``Computer assisted planning for curved laser interstitial thermal therapy,'' \emph{IEEE Transactions on Biomedical Engineering}, vol.~68, no.~10, pp. 2957--2964, 2021.

\bibitem{segato2019automated}
A.~Segato, V.~Pieri, A.~Favaro, M.~Riva, A.~Falini, E.~De~Momi, and A.~Castellano, ``Automated steerable path planning for deep brain stimulation safeguarding fiber tracts and deep gray matter nuclei,'' \emph{Frontiers in Robotics and AI}, vol.~6, p.~70, 2019.

\bibitem{feng2017pose}
M.~Feng, X.~Jin, W.~Tong, X.~Guo, J.~Zhao, and Y.~Fu, ``Pose optimization and port placement for robot-assisted minimally invasive surgery in cholecystectomy,'' \emph{The International Journal of Medical Robotics and Computer Assisted Surgery}, vol.~13, no.~4, p. e1810, 2017.

\bibitem{cannon2003port}
J.~W. Cannon, J.~A. Stoll, S.~D. Selha, P.~E. Dupont, R.~D. Howe, and D.~F. Torchiana, ``Port placement planning in robot-assisted coronary artery bypass,'' \emph{Transactions on Robotics and Automation}, vol.~19, no.~5, pp. 912--917, 2003.

\bibitem{wankhede2019heuristic}
A.~Wankhede, L.~Madiraju, D.~Patel, K.~Cleary, C.~Oluigbo, and R.~Monfaredi, ``Heuristic-based optimal path planning for neurosurgical tumor ablation,'' in \emph{Medical Imaging 2019: Image-Guided Procedures, Robotic Interventions, and Modeling}, vol. 10951.\hskip 1em plus 0.5em minus 0.4em\relax SPIE, 2019, pp. 655--663.

\bibitem{adhami2000planning}
L.~Adhami, E.~Coste-Maniere, and J.-D. Boissonnat, ``Planning and simulation of robotically assisted minimal invasive surgery,'' in \emph{Medical Image Computing and Computer-Assisted Intervention--MICCAI 2000: Third International Conference, Pittsburgh, PA, USA, October 11-14, 2000. Proceedings 3}.\hskip 1em plus 0.5em minus 0.4em\relax Springer, 2000, pp. 624--633.

\bibitem{hayashi2017optimal}
Y.~Hayashi, K.~Misawa, and K.~Mori, ``Optimal port placement planning method for laparoscopic gastrectomy,'' \emph{International Journal of Computer Assisted Radiology and Surgery}, vol.~12, no.~10, pp. 1677--1684, 2017.

\bibitem{sun2007port}
L.~W. Sun and C.~K. Yeung, ``Port placement and pose selection of the {da Vinci} surgical system for collision-free intervention based on performance optimization,'' in \emph{Proc. International Conference on Intelligent Robots and Systems (IROS)}.\hskip 1em plus 0.5em minus 0.4em\relax IEEE, 2007, pp. 1951--1956.

\bibitem{vahrenkamp2016workspace}
N.~Vahrenkamp, H.~Arnst, M.~W{\"a}chter, D.~Schiebener, P.~Sotiropoulos, M.~Kowalik, and T.~Asfour, ``Workspace analysis for planning human-robot interaction tasks,'' in \emph{Proc. International Conference on Humanoid Robots (Humanoids)}.\hskip 1em plus 0.5em minus 0.4em\relax IEEE, 2016, pp. 1298--1303.

\bibitem{mainprice2012sharing}
J.~Mainprice, M.~Gharbi, T.~Sim{\'e}on, and R.~Alami, ``Sharing effort in planning human-robot handover tasks,'' in \emph{Proc. International Symposium on Robot and Human Interactive Communication (RO-MAN)}.\hskip 1em plus 0.5em minus 0.4em\relax IEEE, 2012, pp. 764--770.

\bibitem{walker2015robot}
I.~D. Walker, L.~Mears, R.~S. Mizanoor, R.~Pak, S.~Remy, and Y.~Wang, ``Robot-human handovers based on trust,'' in \emph{2015 Second International Conference on Mathematics and Computers in Sciences and in Industry (MCSI)}.\hskip 1em plus 0.5em minus 0.4em\relax IEEE, 2015, pp. 119--124.

\bibitem{dubins1957curves}
L.~E. Dubins, ``On curves of minimal length with a constraint on average curvature, and with prescribed initial and terminal positions and tangents,'' \emph{American Journal of Mathematics}, vol.~79, no.~3, pp. 497--516, 1957.

\bibitem{cai2017task}
W.~Cai, M.~Zhang, and Y.~R. Zheng, ``Task assignment and path planning for multiple autonomous underwater vehicles using 3d dubins curves,'' \emph{Sensors}, vol.~17, no.~7, p. 1607, 2017.

\bibitem{hota2010optimal}
S.~Hota and D.~Ghose, ``Optimal geometrical path in 3d with curvature constraint,'' in \emph{2010 IEEE/RSJ International Conference on Intelligent Robots and Systems (IROS)}.\hskip 1em plus 0.5em minus 0.4em\relax IEEE, 2010, pp. 113--118.

\bibitem{hota2014optimal}
------, ``Optimal trajectory planning for path convergence in three-dimensional space,'' \emph{Proceedings of the Institution of Mechanical Engineers, Part G: Journal of Aerospace Engineering}, vol. 228, no.~5, pp. 766--780, 2014.

\bibitem{elbanhawi2014randomised}
M.~Elbanhawi and M.~Simic, ``Randomised kinodynamic motion planning for an autonomous vehicle in semi-structured agricultural areas,'' \emph{Biosystems Engineering}, vol. 126, pp. 30--44, 2014.

\bibitem{chitsaz2007time}
H.~Chitsaz and S.~M. LaValle, ``Time-optimal paths for a dubins airplane,'' in \emph{2007 46th IEEE conference on decision and control}.\hskip 1em plus 0.5em minus 0.4em\relax IEEE, 2007, pp. 2379--2384.

\bibitem{fauser2018planning}
J.~Fauser, G.~Sakas, and A.~Mukhopadhyay, ``Planning nonlinear access paths for temporal bone surgery,'' \emph{International journal of computer assisted radiology and surgery}, vol.~13, pp. 637--646, 2018.

\bibitem{choset1999follow}
H.~Choset and W.~Henning, ``A follow-the-leader approach to serpentine robot motion planning,'' \emph{Journal of Aerospace Engineering}, vol.~12, no.~2, pp. 65--73, 1999.

\bibitem{shkel2001classification}
A.~M. Shkel and V.~Lumelsky, ``Classification of the dubins set,'' \emph{Robotics and Autonomous Systems}, vol.~34, no.~4, pp. 179--202, 2001.

\bibitem{ichnowski2018concurrent}
J.~Ichnowski and R.~Alterovitz, ``Concurrent nearest-neighbor searching for parallel sampling-based motion planning in {SO}(3), {SE}(3), and euclidean spaces,'' \emph{Springer}, 2018.

\bibitem{cancerSociety}
{American Cancer Society}, ``Cancer facts and figures,'' \emph{American Cancer Society Tech. Rep.}, 2022.

\bibitem{bray2018global}
F.~Bray, J.~Ferlay, I.~Soerjomataram, R.~L. Siegel, L.~A. Torre, and A.~Jemal, ``Global cancer statistics 2018: Globocan estimates of incidence and mortality worldwide for 36 cancers in 185 countries,'' \emph{CA: a cancer journal for clinicians}, vol.~68, no.~6, pp. 394--424, 2018.

\bibitem{torbenson2015liver}
M.~Torbenson and P.~Schirmacher, ``Liver cancer biopsy--back to the future?!'' \emph{Hepatology}, vol.~61, no.~2, pp. 431--433, 2015.

\bibitem{bodard2023percutaneous}
S.~Bodard, S.~Guinebert, E.~N.~Petre, B.~Marinelli, D.~Sarkar, M.~Barral, and F.~H~Cornelis, ``Percutaneous liver interventions with robotic systems: a systematic review of available clinical solutions,'' \emph{The British Journal of Radiology}, p. 20230620, 2023.

\bibitem{Reed2011_RAM}
{Reed, K. B., Majewicz, A., Kallem, V., Alterovitz, R., Goldberg, K., Cowan, N. J.} and A.~M. Okamura, ``Robot-assisted needle steering,'' \emph{IEEE Robotics and Automation Magazine}, vol.~18, no.~4, pp. 35--46, 2011.

\bibitem{clark2013cancer}
K.~Clark, B.~Vendt, K.~Smith, J.~Freymann, J.~Kirby, P.~Koppel, S.~Moore, S.~Phillips, D.~Maffitt, M.~Pringle \emph{et~al.}, ``The cancer imaging archive (tcia): maintaining and operating a public information repository,'' \emph{Journal of digital imaging}, vol.~26, pp. 1045--1057, 2013.

\bibitem{fried2022clinical}
I.~Fried, J.~A. Akulian, and R.~Alterovitz, ``A clinical dataset for the evaluation of motion planners in medical applications,'' \emph{arXiv preprint arXiv:2210.10834}, 2022.

\bibitem{kikinis20143d}
R.~Kikinis, S.~D. Pieper, and K.~G. Vosburgh, ``3d slicer: a platform for subject-specific image analysis, visualization, and clinical support,'' in \emph{Intraoperative imaging and image-guided therapy}.\hskip 1em plus 0.5em minus 0.4em\relax Springer, 2014, pp. 277--289.

\bibitem{kuntz2016toward}
A.~Kuntz, P.~J. Swaney, A.~Mahoney, R.~H. Feins, Y.~Z. Lee, R.~J. Webster~III, and R.~Alterovitz, ``{Toward transoral peripheral lung access: Steering bronchoscope-deployed needles through porcine lung tissue},'' in \emph{Hamlyn Symposium on Medical Robotics}, 2016, pp. 9--10.

\bibitem{armato2011lung}
S.~G. Armato~III, G.~McLennan, L.~Bidaut, M.~F. McNitt-Gray, C.~R. Meyer, A.~P. Reeves, B.~Zhao, D.~R. Aberle, C.~I. Henschke, E.~A. Hoffman \emph{et~al.}, ``The lung image database consortium ({LIDC}) and image database resource initiative ({IDRI}): a completed reference database of lung nodules on ct scans,'' \emph{Medical physics}, vol.~38, no.~2, pp. 915--931, 2011.

\bibitem{fried2023dataset}
I.~Fried, J.~Hoelscher, J.~A. Akulian, and R.~Alterovitz, ``A dataset of anatomical environments for medical robots: Modeling respiratory deformation,'' \emph{arXiv preprint arXiv:2310.04289}, 2023.

\end{thebibliography}
\newpage
\section{Biography Section}
\begin{IEEEbiography}[{\includegraphics[width=1in,height=1.25in,clip,keepaspectratio]{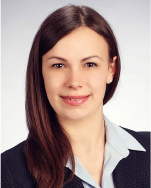}}]{Janine Hoelscher}
(Graduate Student
Member, IEEE) received the B.S. degree in computer science and the M.S. degree in computer
science with a minor in psychology from the
Technical University of Darmstadt, Darmstadt,
Germany, in 2014 and 2017, respectively. She is
currently pursuing the Ph.D. degree in computer
science with the University of North Carolina at
Chapel Hill, Chapel Hill, NC, USA.
Since 2018, she has been a Research Assistant
with the Computational Robotics Research Group, Chapel Hill. Her research
interests include motion planning and machine learning for robotics applications. 
\end{IEEEbiography}

\begin{IEEEbiography}[{\includegraphics[width=1in,height=1.25in,clip,keepaspectratio]{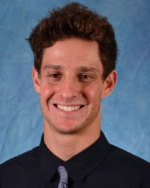}}]{Inbar Fried}
received the B.A. and M.S. degrees in computer science from Tufts University, Medford, MA, USA, in 2014 and 2015, respectively, and the Ph.D. degree in computer science from the University of North Carolina, Chapel Hill, NC, USA in 2023.
He is currently pursuing the M.D. degree with the University of North Carolina School of Medicine as part of the M.D.-Ph.D. combined degree program.
He is currently with the Computational Robotics Research Group.
His research interests include developing motion planning and visual localization algorithms for medical robots to enable novel and improved medical procedures.
\end{IEEEbiography}

\begin{IEEEbiography}[{\includegraphics[width=1in,height=1.25in,clip,keepaspectratio]
{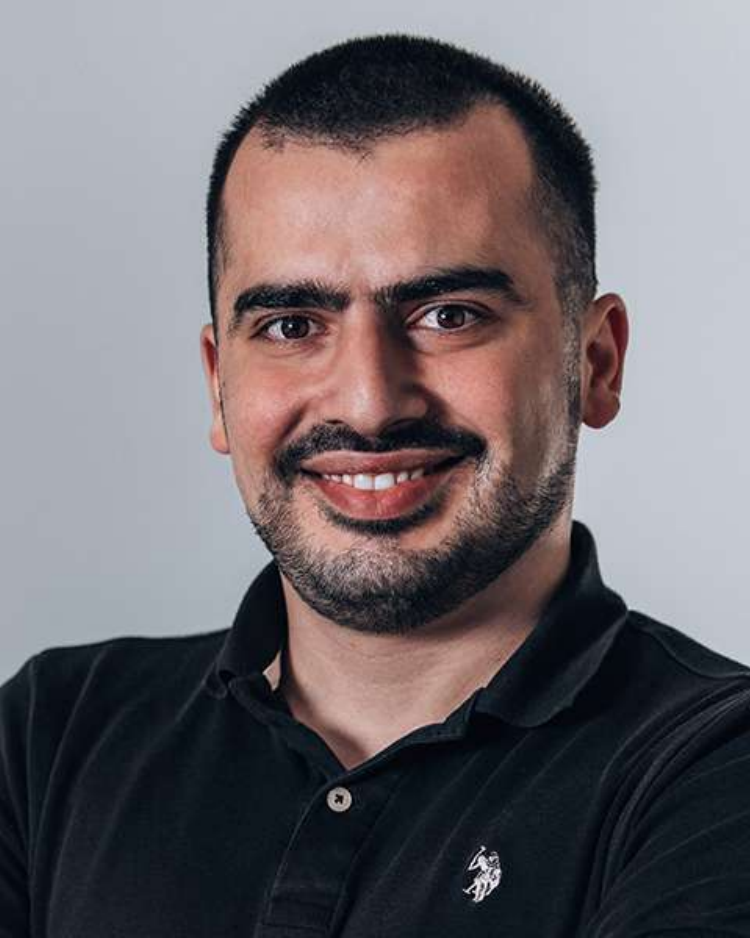}}]{Spiros Tsalikis}
received the B.S. degree in informatics from the Aristotle University of Thessaloniki, Thessaloniki, Greece and the M.S. degree in computer science from the Old Dominion University, Norfolk, VA, USA. He is a R\&D Engineer at Kitware, Inc, and is currently pursuing the Ph.D. degree in computer science with the University of North Carolina at Chapel Hill, Chapel Hill, NC, USA. Since 2023, he has been part of the IRON and Computational Robotics Research groups. His research interests include computational geometry, parallel computing, scientific visualization, graphics and AR/VR.
\end{IEEEbiography}

\begin{IEEEbiography}[{\includegraphics[width=1in,height=1.25in,clip,keepaspectratio]{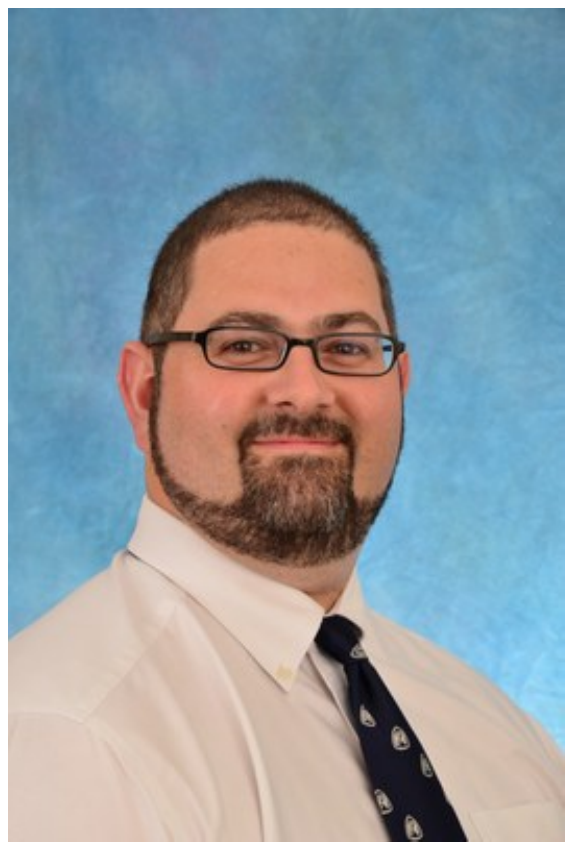}}]{Jason Akulian}
received the B.S. degree in
biochemistry from the University of California,
Riverside, USA, in 1998, the M.P.H. degree in
epidemiology and biostatistics from Loma Linda
University, Loma Linda, CA, USA, in 2001, and
the M.D. degree from St. George’s University,
Grenada, in 2005.
Since 2013, he has been an Assistant Professor of
medicine with the University of North Carolina at Chapel Hill. He is currently
the Director of Interventional Pulmonology and the Carolina Center for
Pleural Disease. His research interests include advanced diagnostic
and therapeutic bronchoscopy and malignant pleural disease.
\end{IEEEbiography}

\begin{IEEEbiography}[{\includegraphics[width=1in,height=1.25in,clip,keepaspectratio]{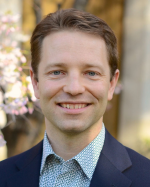}}]{Robert J. Webster III}
(Senior Member,
IEEE) received the B.S. degree in electrical engineering from Clemson University, Clemson, SC,
USA, in 2002, and the M.S. and Ph.D. degrees in
mechanical engineering from Johns Hopkins University, Baltimore, MD, USA, in 2004 and 2007,
respectively.
In 2008, he joined the Faculty Member of Vanderbilt University, Nashville, TN, USA, where he
is currently a Richard A. Schroeder Professor of
mechanical engineering, and also a Professor of electrical engineering, otolaryngology, neurological surgery, urologic surgery, and medicine. His research interests include surgical robotics, image-guided surgery, and continuum robotics. 
\end{IEEEbiography}

\begin{IEEEbiography}[{\includegraphics[width=1in,height=1.25in,clip,keepaspectratio]{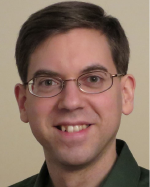}}]{Ron Alterovitz}
(Senior Member, IEEE) received the
B.S. degree (Hons.) from the California Institute
of Technology (Caltech), Pasadena, CA, USA,
in 2001, and the Ph.D. degree in industrial engineering and operations research from the University of California, Berkeley, Berkeley, CA, USA,
in 2006.
In 2009, he joined the Faculty Member of the
Department of Computer Science, University of
North Carolina at Chapel Hill, Chapel Hill, NC,
USA, where he is currently a Professor. His research interests include developing robot motion planning algorithms, with a focus on enabling robots to
perform new, less invasive medical procedures.
\end{IEEEbiography}
\vspace{11pt}

\vfill
\end{document}